\documentclass{article}

\usepackage{jmlr2e}

 \usepackage{enumitem}

\usepackage[american]{babel}

\usepackage[utf8]{inputenc}

\usepackage[colorlinks]{hyperref}

\usepackage{xcolor}
\definecolor{niceblue}{rgb}{0.10, 0.14, 0.76} 

\AtBeginDocument{%
\hypersetup{
    citecolor=niceblue,
    linkcolor=niceblue,   
    urlcolor=niceblue}
}
\hypersetup{citecolor=ultramarine}

\usepackage[T1]{fontenc}    
\usepackage{booktabs}       
\usepackage{amsfonts}       
\usepackage{nicefrac}       
\usepackage{microtype}      
\usepackage{amssymb}
\usepackage{amsmath}
\usepackage{amsthm}
\usepackage{xspace}
\usepackage{wrapfig}

\usepackage[%
    minnames=1,maxnames=99,maxcitenames=3,
    style=numeric-comp,
    sorting=none,
    doi=false,url=false,
    giveninits=true,
    hyperref,natbib,backend=biber]{biblatex}
\renewbibmacro{in:}{%
  \ifentrytype{article}{}{\printtext{\bibstring{in}\intitlepunct}}}
\bibliography{sample}

\usepackage{graphicx}
\usepackage{colordvi}
\usepackage{mathtools}
\usepackage{subcaption}

\usepackage[capitalize]{cleveref}

\usepackage[ruled,vlined]{algorithm2e}

\hypersetup{citecolor=blue}

\usepackage[hide=false,setmargin=true,marginparwidth=1.25in]{marginalia}
\newtheorem{thm}{Theorem}
\newtheorem{lemma}{Lemma}

\newcommand{\EE}{\mathbb{E}}
\newcommand{\Drift}{\Delta_i}

\newcommand{\iid}{i.i.d.\xspace}

\newcommand{\DataPrior}{S_P}
\newcommand{\Dist}{\mathcal D}

\newcommand{\detmask}{b}

\newcommand{\KLname}{\mathrm{KL}}
\newcommand{\KL}[2]{\KLname(#1 \,\|\,#2)}
\newcommand{\klbin}[2]{\mathrm{kl}(#1 \,\|\,#2)}
\newcommand{\klinv}[2]{\mathrm{kl}^{-1}\left (#1 \,\|\,#2 \right)}

\newcommand\optparen[1]{\ifthenelse{\equal{#1}{}}{}{(#1)}}
\newcommand{\loss}{\ell}
\newcommand{\surloss}{\breve\loss}

\newcommand{\RiskChar}{R}
\newcommand{\Risk}[2]{\RiskChar_{#1}\optparen{#2}}
\newcommand{\EmpRisk}[2]{\hat \RiskChar_{#1}\optparen{#2}}

\newcommand{\SurEmpRisk}[2]{\tilde \RiskChar_{#1}\optparen{#2}}

\newcommand{\priorpruneprob}{\lambda^{0}}
\newcommand{\hatpriorpruneprob}{\hat{\lambda}^{0}}

\newcommand{\SbarP}{S_{\smash{\bar  P}}}

\newcommand{\features}[1]{\psi(#1)}
\newcommand{\featurei}[2]{\psi_{#2}(#1)}

\newcommand{\Reals}{\mathbb{R}}

\newcommand{\NNReals}{\Reals_+}
\newcommand{\E}{\mathbb{E}}

\newcommand{\sign}{\mathrm{sign}}

\newcommand{\weight}{w^P}

\newcommand{\ent}{\textrm{Ent}}

\newcommand{\defn}[1]{\emph{#1}}

\newcommand{\norm}[1]{\|#1\|_2}
\newcommand{\Transpose}[1]{#1^{\mathrm{T}}}

\newcommand{\rweights}{W}
\newcommand{\mweights}{\overline{W}}
\newcommand{\Trace}[1]{\mathrm{Tr(#1)}}

\newcommand{\NormalD}[2]{\mathcal{N}(#1,#2)}

\begin{document}
\title{Probabilistic fine-tuning of pruning masks \\and PAC-Bayes self-bounded learning} 

\author{%
  Soufiane Hayou\thanks{Correspondence to: \texttt{soufiane.hayou@yahoo.fr} and \texttt{gkdz@google.com}}\\
  \addr{Department of Statistics}\\
  \addr{University of Oxford}\\
  \AND
  Bobby He \\
  \addr{Department of Statistics}\\
  \addr{University of Oxford}\\
  \AND
  Gintare Karolina Dziugaite\footnotemark[1]\ \thanks{This work was carried out while the author was at ServiceNow. It was finalized at Google Brain.}\\
  \addr{MILA}
}

\editor{}
\maketitle

\begin{abstract}
We study an approach to learning pruning masks by optimizing the expected loss of stochastic pruning masks, i.e., masks which zero out each weight independently with some weight-specific probability. We analyze the training dynamics of the induced stochastic predictor in the setting of linear regression, and observe a data-adaptive $L_1$ regularization term, in contrast to the data-adaptive $L_2$ regularization term known to underlie dropout in linear regression.
We also observe a preference to prune weights that are less well-aligned with the data labels. We evaluate probabilistic fine-tuning for optimizing stochastic pruning masks for neural networks, starting from masks produced by several baselines (namely, magnitude pruning \citep{han2015learning}, SNIP \citep{lee2018snip}, and random masks). In each case, we see improvements in test error over baselines, even after we threshold fine-tuned stochastic pruning masks. Finally, since a stochastic pruning mask induces a stochastic neural network, we consider training the weights and/or pruning probabilities simultaneously to minimize a PAC-Bayes bound on generalization error. Using data-dependent priors \citep{dziugaite2020role}, we obtain a self-bounded learning algorithm with strong performance and numerically tight bounds. In the linear model, we show that a PAC-Bayes generalization error bound is controlled by the magnitude of the change in feature alignment between the ``prior'' and ``posterior'' data.

\end{abstract}
\section{Introduction}
\label{sec:intro}
Modern deep neural networks (DNNs) are
massively overparametrized, which presents problems both during training and deployment. Recently, the demand to reduce the size of networks during and after training has
led to a rush of work in network pruning (\citealt{han2015learning, frankle2018lottery, lee2018snip, wangGRASP, hayouPruning}, etc.), 
much of it building on pioneering efforts carried out decades ago  \citep{mozer_pruning,lecun_pruning,hassibi_pruning}. Many approaches to pruning begin by masking a subset of weights to zero. In approaches applied after training, this step typically causes a drop in both train and test accuracy, which is then  addressed by fine-tuning (or retraining).

Intuitively, fine-tuning can be seen as looking for \emph{locally optimal weights} given the mask.
In this work, we consider replacing adhoc rules for selecting pruning masks with optimization to learn the masks. 
In doing so, we observe that the masks used by state-of-the-art approaches to pruning are far from locally optimal.

To approach this learning problem, we introduce a stochastic optimization framework for masks, based on the Gumbel--Softmax trick \citep{Jang2017,Madd2017}.
Rather than viewing the application of a mask as a deterministic procedure,
we view it as a stochastic operation, where each weight $w_i$ is pruned with some probability $1-\lambda_i$ 
such that $D-\sum_i \lambda_i = sD$, where $D$ is the total number of weights and $s \in [0,1]$ is the desired sparsity level.
We then consider optimizing  
the \emph{expected} empirical risk of this induced randomized predictor
with respect to the pruning probabilities. 
Finally, to obtain a deterministic mask, we threshold the probabilities at a fixed level or at a level chosen to achieve a desired sparsity.
We call this entire procedure \emph{probabilistic fine-tuning} (PFT).

We investigate probabilistic fine-tuning both analytically and empirically. 
To gain some analytical insight, we study the case of linear regression in \cref{sec:linmodelanalysis}, where we find that optimizing the pruning probabilities of a stochastic pruning mask has a data-dependent regularizing effect similar to that of dropout \citep{srivastava2014dropout}.
For linear regression, it is well known that dropout leads to data-adaptive $L_2$ regularization of the weights \citep{WagerWangLiang2013}. We show that optimizing the pruning probabilities looks like training the mean network subject to data-and-weight-adaptive $L_1$ regularization of the pruning probabilities. In the appendix, we discuss how this analysis of a linear model can be extended to wide DNNs in the Neural Tangent Kernel regime \citep{jacot}.

We evaluate probabilistic fine-tuning on several benchmark vision tasks
in \cref{sec:pftexperiments,sec:pbpexperiments}. In this setting, the regularizing effects of dropout on weight training are well documented. On the other hand, the dynamics of optimizing only the dropout probabilities has not received close attention,
although the sparsifying effects of variational dropout (when both weights and Gaussian dropout parameters are trained) were noted by \citet{molchanov2017variational}.
We consider optimizing stochastic pruning masks, starting from initial pruning probabilities derived from
magnitude pruning \citep{han2015learning}, SNIP \citep{lee2018snip}, and random pruning. %
Across the board, we observe improved accuracy as a result of probabilistic fine-tuning, demonstrating that existing heuristics are, in some cases, far from optimal.

Finally, after observing that pruned networks are also more robust to random perturbations of their weights, we consider replacing deterministic fine-tuning with PAC-Bayes bound optimization.
Our approach, dubbed PAC-Bayes pruning, uses spike-and-slab prior and posterior distributions, which were previously investigated by \citet{smart15} in the context of variational inference.
Different from earlier work, we employ data-dependent priors to achieve tighter bounds (and prevent underfitting), as advocated recently by \citet{dziugaite2020role}.
On MNIST \citep{lecun-mnisthandwrittendigit-2010}, Fashion-MNIST \citep{xiao2017fashion}, and CIFAR-10 \citep{Krizhevsky09learningmultiple}, we obtain competitive predictors along with numerically tight, nonvacuous generalization bounds. 

\paragraph{Contributions}
We propose to learn pruning masks by truncating stochastic pruning masks that have been optimized to minimize expected error using stochastic optimization.
We study this idea analytically and empirically and describe several extesions.

\begin{itemize}[leftmargin=1.5em]
    \item We introduce \emph{Probabilistic Fine-Tuning} (PFT), whereby we learn a pruning mask for a trained predictor. 
    \item Theoretically, PFT trains a mean predictor subject to implicit data-and-weight dependent L1 regularization.
     Empirically, 
    PFT improves upon popular heuristics for choosing pruning masks, demonstrating these approaches are far from locally optimal.

    \item Based on the observation that pruned networks are more robust to perturbations of the weights, we propose \emph{PAC-Bayes Pruning} (PBP), whereby we prune a neural network by optimizing a PAC-Bayes bound, using data-dependent, spike-and-slab priors and posteriors on the weights.
    \item Theoretically, PBP performs constrained feature realignment in linear models. Empirically, it delivers strong predictors for standard benchmarks along with numerically tight generalization bounds.
\end{itemize}

\section{Stochastic pruning}
\label{sec:probpruning}
Consider a class of predictors $\{f_W\}$, indexed by weight vectors $W \in \Reals^D$.
Let $S$ 
denote the training data,
consisting of $N$ i.i.d.\ samples from an unknown distribution $\Dist$. 
For a given loss function $\ell$,
let $\EmpRisk{S}{W}$ denote the empirical risk, i.e.,
the loss
of the predictor $f_W$ on average over the training set $S$.
Let $\Risk{\Dist}{W}$ denote the risk, i.e., the expectation of the empirical risk, regarding the weights as fixed.

Pruning involves the application of a binary mask $\detmask \in \{0,1\}^D$
to the weights of a predictor $f_W$ producing the predictor $f_{W'}$, where $W'= \detmask \circ W$ is the Hadamard (i.e.,  element-wise) product.
We say weight $i \in [D]$ is pruned if $b_i = 0$.
We assume our class of predictors $f_W$ are such that $b_i=0$ implies $\frac{\partial f_{W'}(x)}{\partial w_i} = 0$ for all $x \in X$. That is, a pruned weight plays no role in training.

\paragraph{Pruning Masks}
A standard approach to generating masks is to give each weight $w_i$ a score $g_i$ according to some criterion. 
The mask is then created by keeping the top $k$ weights by score, where $k$ is chosen to meet some desired level of sparsity.
Examples of pruning criterion include: 
\begin{itemize}[leftmargin=1.5em]
    \item {\bf Magnitude}: weights are scored by magnitude, i.e., $g_i = |W_i|$ \citep{han2015learning};
    \item {\bf Sensitivity}: weights are scored based on how much zero'ing them will impact empirical risk \citep{lecun_pruning, lee2018snip}.
         For example, SNIP \citep{lee2018snip} uses a first-order approximation of the training objective
              $\tilde{\loss}_{W} \approx \tilde{\loss}_{W=0} + W \frac{\partial \tilde{\loss}}{\partial W}$,
         yielding the score $g_i = |W_i \frac{\partial \tilde{\loss}}{\partial W_i}|$.  
\end{itemize}
\paragraph{Stochastic Pruning Masks}
In this work, we consider \emph{stochastic} pruning masks, i.e.,
vectors $\detmask$ of Bernoulli random variables.
We focus on the case where elements of the mask are independent, so that the distribution
of the mask is determined by probabilities $\lambda_i = \Pr\{\detmask_i = 1\}$. 
For a fixed set of weights $W$,
a stochastic mask induces a distribution, say $P$, on pruned weights 
$W'=(w'_i)_{i=1}^D$. 
In particular, conditional on $W$, the components of $W'$ are independent and $w'_i$ has distribution $(1-\lambda_i) \delta_0 + \lambda_i \delta_{w_i}$, where $\delta_v$ denotes the Dirac mass on $v$.
\paragraph{Optimizing stochastic pruning masks}
Besides inducing a distribution $P$ on weights, stochastic pruning masks correspond to randomized predictors:
We can think of such a distribution as a stochastic predictor that, on each input $x$, samples $W \sim P$ and then outputs $f_W(x)$. Such a stochastic predictor is known as a \emph{Gibbs} predictor.
The \defn{empirical Gibbs risk} 
and \defn{Gibbs risk} of $P$ are, respectively,
$
\EmpRisk{S}{P} = \E_{W \sim P}[\EmpRisk{S}{W}]$
and
$\Risk{\Dist}{P} = \E_{W \sim P}[\Risk{\Dist}{W}].
$
 
We can find stochastic pruning masks that produce sparse networks with small empirical risk by minimizing the empirical Gibbs risk with respect to the pruning probabilities.
A discussion on how we initialize the pruning probabilities is provided in \cref{app:initializing_pruning_probs}.

\subsection{Probabilistic Fine-Tuning (PFT)}
Here we outline the steps of the the PFT algorithm that we use to optimize pruning masks:
\paragraph{Stage 1: Pre-train the dense model.}
Using a variant of stochastic gradient descent, minimize $\EmpRisk{S}{W}$.
Obtain a deterministic predictor with weights $\hat{W}$.
\paragraph{Stage 2: Fine-tune the masks.}
\begin{enumerate}
    \item Initialize the pruning probabilities.
    If a criterion $g$ is used, evaluate it on $\hat{W}$.
    \item Continue minimizing the empirical Gibbs risk.
    The minimization is done w.r.t. the pruning probabilities (performed using the Gumbel--Softmax trick. See \cref{app:experimentaldetails}) and the weights.
    \item The trained pruning probabilities are used to create a new \emph{deterministic} mask by keeping the weights with the largest $\lambda_i$ values. \footnote{Note that optimizing directly over the mask is computationally infeasible. Our PFT algorithm is a stochastic optimization method over the mask.} The threshold is chosen based on the desired sparsity level.
    \item Fine-tune the weights of the sparse model.
\end{enumerate}

\section{Linear model analysis}
\label{sec:linmodelanalysis}
In this section, we present a theoretical analysis of PFT in the case of linear models. 
Let $S_P$ denote the dataset we use to train the stochastic predictor $P$. Let $w^P$ denotes the trained weights. 
Write $M=|S_P|$.
Let $X_P \in \Reals^{M \times d}$ denote the matrix of inputs and $Y_P \in \Reals^{M \times 1}$ the vector of outputs.
Let $\psi: \Reals^d \mapsto \Reals^D$ be a zero-mean feature map, 
and write $\featurei{X_P}{i} \in \Reals^{M}$ for the vector of the values of the $i^{\mathrm{th}}$ feature, i.e., the vector whose $j^{\mathrm{th}}$ entry is $\featurei{x_j}{i}$ for $x_j \in X_P$. 
For the empirical covariance of the features, write $\Sigma_P = \frac{1}{M} \sum_{x \in X_P} \features{x} \features{x}^T$. In this section, we study a linear predictor $f(x){=}\features{x}^T w$, with weights $w$, trained using squared error loss 
$\loss(f(x),y) = (f(x) - y)^2/2$.

Consider a two-stage training procedure: 

\begin{enumerate}
    \item Train the linear model with $(X_{P}, Y_{P})$. 
    Let $\weight$ denote the trained weights.
    \item Learn pruning probabilities $\{ \priorpruneprob_i\}_{i=1}^D$ by minimizing $\EmpRisk{S}{P}$, where $P$ is a product distribution $\prod_{i=1}^D p_i$, with $p_i = (1-\priorpruneprob_i) \delta_{\{0\}} + \priorpruneprob_i \delta_{\{\weight_i\}}$.

\end{enumerate}

Letting $\rweights {\sim} P$,
the empirical risk of $P$ under square loss 
satisfies
\begin{gather}
\begin{split}
2 \EmpRisk{S_{P}}{P} 
= \frac 1 M \EE [ \norm{ \features{X_{P}}^T \rweights - Y_{P} }^2 ] \hphantom{WWWW}
\\
= \EE[ \Transpose{\rweights} \Sigma_P \rweights] - 2 \Transpose{\mweights} A(P) + \frac 1 M \norm{Y_{P}}^2,
\end{split}
\label{eq:emprisklinearmodel}
\end{gather}
where $\mweights$ is the mean of $\rweights$ 
with entries $\mweights_i = \priorpruneprob_i \weight_i$
and $A(P)$ is the \emph{feature alignment vector} for the data $S_P$,
with entries
$A_i(P) = \frac{1}{|X_P|}  \featurei{X_P}{i}^T Y_{P} $, 
i.e., $A_i(P)$ measures how closely feature $i$ is aligned to the outputs $Y_P$ \citep{elesedy2020lottery}.
Then $\EE[ \Transpose{\rweights} \Sigma_P \rweights] = \Trace{\Gamma \Sigma_P}$, where
\begin{equation}
    \Gamma_{i,j} =
    \begin{cases*}
      \priorpruneprob_i \priorpruneprob_j \weight_i \weight_j, & if $i \neq j$, \\
      \priorpruneprob_i (\weight_i)^2,        & otherwise.
    \end{cases*}
\end{equation}
Our analysis focuses on $\frac{\partial \EmpRisk{S_{P}}{P}}{\partial \priorpruneprob_i}$,
the derivative of \cref{eq:emprisklinearmodel} with respect to $\priorpruneprob_i$ given by,
\begin{equation}
 \sum_{i \neq j} \priorpruneprob_j \weight_i \weight_j (\Sigma_{P})_{ij} + \frac 1 2 (\weight_i)^2 (\Sigma_{P})_{ii}
-  \weight_i A_i(P).
      \label{eq:derivativeofemprisklinear}
\end{equation}

\subsection{Learning prior pruning probabilities.}
\label{sec:learning_prior_probs}

We now investigate the effect of optimizing the empirical risk of the stochastic predictor $P$ with respect to $\priorpruneprob_i$'s, as one does in stage 2 of PFT.
According to the following result, this optimization is  equivalent to optimizing the empirical risk of the mean predictor $\mweights$ with an explicit regularization term that encourages pruning.
(See \cref{app:learning_prior_probs} for a formal statement and proof.)

\begin{lemma}[Informal]
Gradient descent on the empirical Gibbs error (\cref{eq:emprisklinearmodel})
with respect to pruning probabilities $\{\lambda_i\}_{i=1}^D$ performs implicit $L_1$ regularization of $\{\lambda_i\}_{i=1}^D$.
\label{lemma1:statement}
\end{lemma}

\vspace*{-1em}
\begin{proof}[Proof sketch]
One can show $  \frac{\partial \EmpRisk{S_P}{P}}{\partial \priorpruneprob_i} 
    {-} \frac{\partial\EmpRisk{S_P}{\mweights}}{\partial \priorpruneprob_i} 
    {=} r_i^0,$
where $r_i^0 = \frac 1 2 (\weight_i)^2 (\Sigma_{P})_{ii}$.
For fixed weights $\weight_i$, 
$r_i^0$ 
is label-independent and can be interpreted as a stochastic pruning regularizer, pushing $\priorpruneprob_i$ closer to 0 (i.e., $r^0_i$ induces an $L_1$ penalty on $\lambda^0_i$). 
\end{proof}
Unlike optimizing with respect to the deterministic pruning weight, optimizing with respect to the stochastic pruning mask accounts for the effect of dropping the weight.
While we do not optimize the weights in Stage 2 of the simplified algorithm above, we can apply a similar analysis as in \cref{lemma1:statement} and compare the gradient with respect to the weight $\weight_i$, when differentiating a stochastic predictor versus the mean predictor.
We find
$
    \frac{\partial \EmpRisk{S_P}{P}}
        {\partial \weight_i}  = 
            \priorpruneprob_i \frac{\partial \EmpRisk{S}{\mweights}}
                {\partial \mweights_i} 
                    +\priorpruneprob_i \weight_i (\Sigma_{P})_{ii}.
$ 
In comparison to 
the gradient with respect to mean weights,
$ \frac{\partial \EmpRisk{S}{\mweights} }{\partial \mweights_i} $,
we observe (i) weights that are unlikely to be pruned ($\lambda_i\approx 1$) have a larger effective learning rate; and (ii) optimizing the Gibbs risk introduces a label-independent regularizer $\bar{r}_i^0 = \priorpruneprob_i \weight_i (\Sigma_{P})_{ii}$, which drives small $\lambda_i$ to zero,
in contrast to $L_2$ regularization.
The $L_2$ regularization effect of dropout in linear models is well documented \citep{WagerWangLiang2013,wang2013fast}. 

The regularization terms $\bar{r}_i^0$ and $r_i^0$
appearing in the gradient of empirical Gibbs risk arise due to the term $\EE[ \Transpose{\rweights} \Sigma_P \rweights]$ in \cref{eq:emprisklinearmodel}. Note the interaction between the gradient updates 
with respect to $\lambda_i$ and $\weight_i$:
as $\lambda_i$ increases, $\Big|\frac{ \partial \EmpRisk{S_P}{P}}{\partial \weight_i} \Big|$ grows, thus increasing the magnitude of $\weight_i$, 
and vice versa.

\paragraph{Pruning probabilities are aligned with weight magnitude growth.} In \cref{subsec:weightmag}, we show that in the simple setting of a diagonal covariance matrix, early stopping induces a pruning criterion that is aligned with the magnitude of the weights. In linear regression with diagonal feature covariance, the induced pruning criterion is based on variance-normalized feature alignment.
We further extend the analysis when the features are correlated.

\section{Experiments: Stochastic Pruning}
\label{sec:pftexperiments}

We now experimentally verify the characteristics and performance of stochastic pruning, with a focus on using PFT to fine-tune the masks obtained by popular heuristics. 
In interest of space, full details on the networks, datasets,  hyperparameters, optimizers, and loss functions are provided in \cref{app:experimentaldetails}. 
\cref{app:additionalresults} contains a number of additional experiments on PFT.

\paragraph{Probabilistic fine-tuning boosts the accuracy of sparse predictors}
\label{sec:PFT}
\begin{figure}[t]
    \centering
    \includegraphics[width=0.8\linewidth]{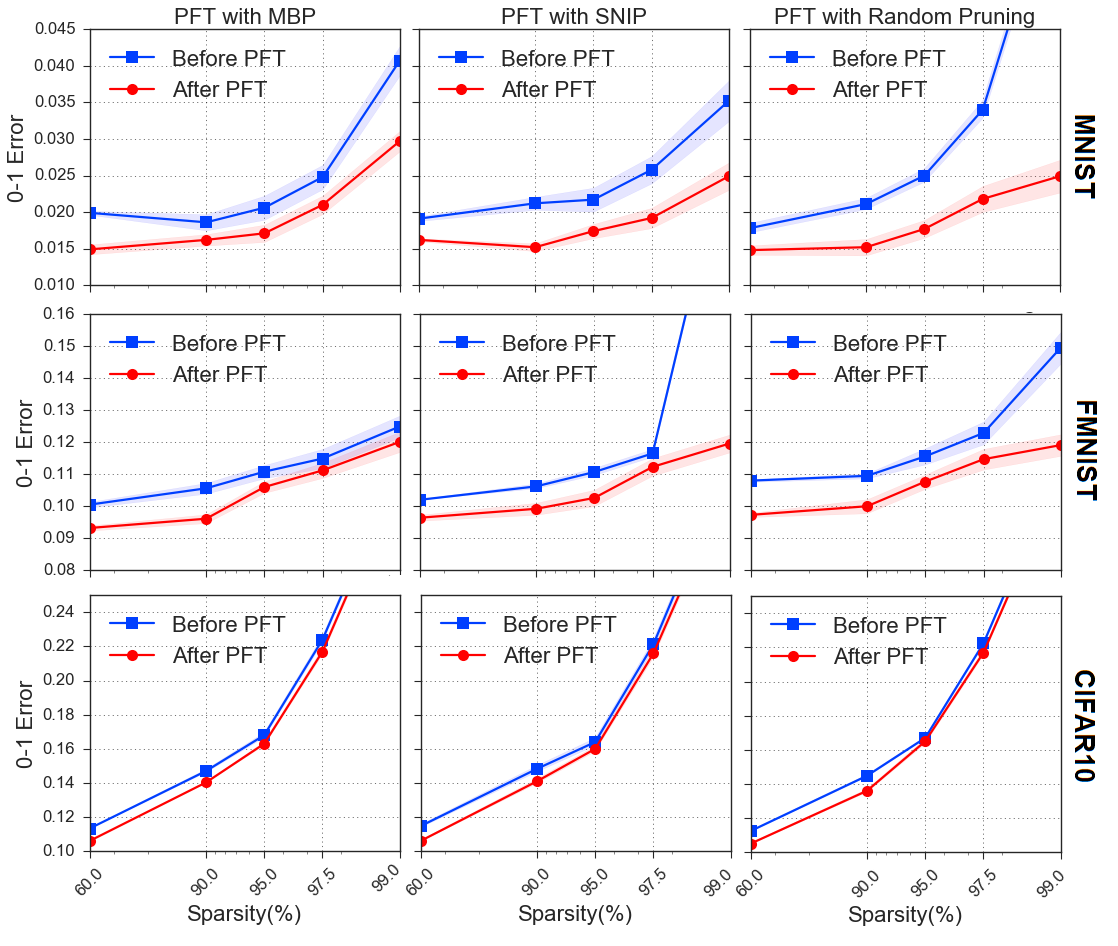}
    \caption{\small{Impact of the application of PFT on standard One-Shot Pruning methods (MBP, SNIP) and Random Pruning. MNIST and Fashion-MNIST with a 3-layer width-1000 MLP and CIFAR-10 with ResNet20.}}
    \label{fig:pft_results}
\end{figure}

One-Shot Pruning (OSP) methods create the mask in a single or no pass over the dataset. Popular choices include pruning at initialization methods \citep{lee2018snip, wangGRASP, hayouPruning, tanaka2020pruning}. 
OSP later in training was discussed in \citep{frankle2020missing}. These methods follow the same protocol: create a mask some time during training (e.g., at initialization or after training), apply the mask to the network, and train the sparse network. PFT aims to improve the performance of OSP methods using stochastic models to fine-tune the mask before training the sparse network. 
This entails training the prior probabilities as discussed in \cref{sec:learning_prior_probs} in the context of linear models; we then create a new mask by keeping the weights with the highest (trained) probabilities and pruning the rest.

\cref{fig:pft_results} shows that PFT consistently boosts the performance of standard OSP methods on small-scale vision tasks; in general, the benefit of applying PFT
is most noticeable for high sparsity levels (e.g., $99\%$) where some OSP methods alone fail to achieve good test error. On CIFAR-10 with ResNet20, PFT improves the accuracy of OSPs by $\sim 1\%$ on average; results on CIFAR-10 with sparsity level $99\%$ are included in \cref{app:additionalresults}.
Note that PFT with Random Pruning (RP) is comparable to other methods, hence suggesting that PFT alone (preceded by RP) can serve as a standalone pruning method. More details and experiments on PFT are provided in \cref{app:pft}.

\begin{figure}
 \centering
    \includegraphics[width=.43\columnwidth]{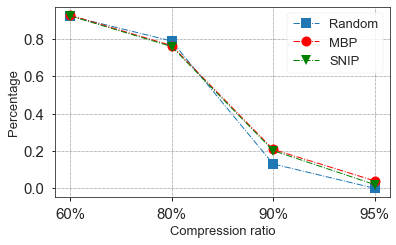}
   \includegraphics[width=.43\linewidth]{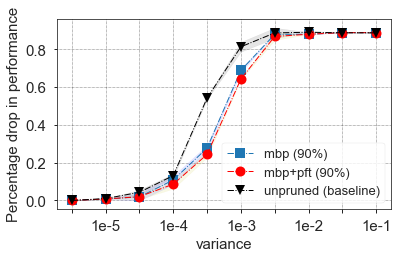}
    \caption{
    \emph{Left}: \small{Percentage of shared weights between OSP masks and masks obtained after applying PFT (CIFAR10 with ResNet20). This shows that OSP methods tend to miss the local optimal masks more and more as we increase the sparsity.}
    \emph{Right}: \small{Drop in performance (as a fraction) of probabilistic model as a function of variance (unpruned versus 90\% ResNet20.)}
    }
    \label{fig:pct_shared_weights}
\end{figure}

\paragraph{As sparsity increases, one-shot pruning (OSP) masks overlap less with fine-tuned ones}
Knowing that masks created by OSP methods are not locally optimal, we intuitively expect that as we increase the sparsity, OSP methods will `miss' the local optimal mask by a growing `distance'. 
We measure this distance by the percentage of weights shared between OSP masks and the masks created after applying PFT. \cref{fig:pct_shared_weights} (left) shows this percentage as a function of sparsity (compression ratio) for different OSP methods (Random, MBP, SNIP) on CIFAR10 with ResNet20 architecture. For a medium range sparsity ($60\%$), around $93\%$ of the weights are shared between masks before and after applying PFT, suggesting that masks found by PFT are close to the original OSP masks. However, this percentage of shared weights between masks before 
and after PFT follows a decreasing pattern w.r.t the sparsity. For high sparsities, e.g., $95\%$ (average over 5 runs), we observe that for MBP and SNIP, only a fraction $\sim 3\%$ of the weights is shared, and for random pruning, this fraction is $0\%$ in all of the runs. This confirms our intuition that the higher the sparsity, the worse the OSP mask, and suggests that PFT searches `farther' for better masks.
\paragraph{Robustness of pruned vs unpruned nets} 

Stochastic masks define Gibbs predictors/classifiers. It is interesting to consider 
how stable the training error of these classifiers are to perturbations also of the underlying weights, which would be indicative of the classifiers having small generalization error \citep{mcallester1999pacbayes,langford2001pacbayes,dziugaite2020role}.
We empirically evaluate whether standard pruned models are more robust to weight perturbations.
\cref{fig:pct_shared_weights} (right) shows the percentage drop in performance of a deterministic trained model (pruned and unpruned) when the weights are perturbed with a Gaussian noise with variance $v^2$, i.e., $W_i \to W_i + v \zeta_i$ for all $i$ where $(\zeta_i)$ are \iid Gaussian variables with distribution $\NormalD{0}{1}$. 
In the variance range where the drop in performance is most noticeable, the pruned network tends to be more robust to perturbations compared to the full network. 
This can also be interpreted as follows: the variance level $v$ that corresponds to a given drop in performance for the pruned model is higher than that which corresponds to the full network with the same drop in performance. In addition, PFT seems to find more robust masks than the original mask (In this example, the one we obtain with MBP). We believe the implicit regularization underlying probabilistic fine-tuning increases this notion of robustness. We explore the implications of this for a PAC-Bayesian analysis in the next section.
\section{PAC-Bayes: Preliminaries}
In the previous section, we saw that pruned networks were more robust to weight perturbation, and that this effect was true also for probabilistically fine-tuned masks. This insensitivity-to-weight-perturbation is suggestive that an application of the PAC-Bayes theorem may yield a generalization bound. 
In this section, we introduce data-dependent PAC-Bayes bounds and propose to optimize them to prune and fine-tune the network. 
We extend our linear model analysis to this approach and find that it carries out constrained feature realignment. We then evaluate this approach empirically, and find that it produces predictors with both high accuracy and extremely tight generalization guarantees, without the need for extra data.

\subsection{Data-dependent PAC-Bayes Bounds}
PAC-Bayes theorems provide 
confidence intervals for Gibbs risks $\Risk{\Dist}{Q}$ in terms of an empirical Gibbs risk $\EmpRisk{S}{Q}$
and the Kullback--Liebler (KL) divergence between any distribution $Q$, and a distribution $P$, which cannot depend on all of the training data $S$.
In PAC-Bayes parlance, $P$ is called a ``prior'' and each $Q$ is referred to as a ``posterior'', on the account that the bound applies also to distributions $Q$ that may depend on the training data $S$.

Many forms of PAC-Bayes theorems exist.
In our work we consider a PAC-Bayes bound with a data-dependent prior, meaning that the prior \emph{depends on} a subset of the training data, $S_P \subset S$, $S$ being the full dataset.
Let $\alpha = |S_P|/|S| \in [0,1)$. In this work we typically use $\alpha\in[0.4,0.8]$. In our experimental setting, we optimize through the following bound due to \citep[][Thm.~D.2]{dziugaite2020role} and \citep[][Sec.~1.3]{catoni2007pacbayes}
\footnote{See \citep[][Fig.~6]{dziugaite2020role} for the discussion on the tightness of the bounds in different regimes.}.

\begin{thm}
\label{thm:pac_bayes_data_dependent}
For every $[0, 1]$-valued loss function $\ell$, 
data distribution $\Dist$, natural $N \in \mathbb{N}$, probability $\delta \in (0, 1)$, with probability at least $1-\delta$ over training data $S \sim \Dist^N$ and all distributions $P$ and $Q$ on weights, where $P$ may only depend on $S_{P} \subset S$,
\begin{equation}
\Risk{\Dist}{Q} \leq \EmpRisk{S\setminus S_P}{Q} + \min
\begin{cases} \varepsilon + \sqrt{ \varepsilon(\varepsilon+2 \EmpRisk{S \setminus S_P}{Q}) }   ,\\
 \sqrt{ \varepsilon/ 2 }   ,
\end{cases}
    \label{eq:datadeppacbayesbound}
\end{equation}
where $\varepsilon = \frac{ \KL{Q}{P}  + \log \frac{2 \sqrt{ (1-\alpha) |S| } }{\delta}}{(1-\alpha) |S|}$.
\end{thm}

As one may expect, larger $\alpha$ yields a better prior $P$ with a smaller $\KL{Q}{P}$. However, the KL term in \cref{eq:datadeppacbayesbound} is divided by $(1-\alpha) |S|$, and increasing $\alpha$ may lead to a looser bound. Therefore, the final risk bound trades-off the decrease in $\KL{Q(S)}{P(S_P)}$ and the change in
$\frac{\KL{Q(S)}{P(S_P)}}{(1-\alpha) |S|}$.

Data-dependent PAC-Bayes bounds previously appeared in many articles, usually as an attempt to empirically approximate distribution-dependent bounds (e.g., \citep{catoni2007pacbayes,lever2010distribution,Amb07,negrea2019information,haghifam2020sharpened,Oneto2016,Oneto2017,DR18,rivasplata2018pac,awasthi2020pac,dziugaite2020role}).

Informally, in the context of stochastic pruning, t\cref{thm:pac_bayes_data_dependent} says that if we do not change the pruning probabilities too much as we get more data, i.e., from $P$ to $Q$, then we will have a small generalization gap.

\paragraph{Optimizing PAC-Bayes bounds.}
In light of \cref{thm:pac_bayes_data_dependent}, 
one approach to learning is to minimize the upper bound on the right-hand side in terms of $Q$.\footnote{Because the KL divergence is finite only if $Q$ is absolutely continuous with respect to $P$, this approach to learning produces a stochastic predictor.} In this work, we consider the application of PAC-Bayes bound optimization to optimizing stochastic pruning masks and fine-tuning the weights.

While for certain forms of PAC-Bayes bounds there exist analytical solutions for the optimal $Q$, the resulting predictors are generally intractable. Instead, one typically performs explicit optimization over a parameterized family of posteriors. A common choice, which can be justified on theoretical grounds asymptotically \citep{alquier2015pacbayes},
is to optimize $Q$ over the class of Gaussian distributions, and choose the prior $P$ to be Gaussian as well. 
For computational reasons, it is common to restrict attention to diagonal covariance for neural networks, following \citet{dziugaite2017computing}.\\

\paragraph{The prior $P$ and posterior $Q$.} 
We consider a spike-and-slab prior $P$ whose slab is a multivariate Gaussian
$P^0$ with mean weight vector $W^0 \in \Reals^D$ and diagonal covariance with the $i^{\mathrm{th}}$ entry $(\sigma_i^0)^2 \in \Reals_{>0}$.
That is, the weights are a priori independent and each weight $w_i$ has a distribution
\begin{align}\label{eq:prior}
p_i = (1-\priorpruneprob_i) \delta_{\{0\}} + \priorpruneprob_i \NormalD{W^0_i}{(\sigma_i^0)^2}.
\end{align}

In order to have a tractable KL divergence, we also optimize over the family of a spike-and-slab posteriors 
whose slabs are multivariate Gaussians $Q^f$ with mean weight vector $W^f$ and diagonal covariance with entries $(\sigma_i^f)^2$. We denote the marginal posteriors by $q_i$, in resemblance to $p_i$, and the posterior probabilities by $\lambda_i$. 

The parameters $\lambda_i$, $W_i^f$, and $\sigma_i^f$ can be optimized while leaving the PAC-Bayes bound valid.
\paragraph{Kullback--Leibler divergence.}
With $P$ and $Q$ as above, $\KL{Q}{P} = \sum_{i=1}^D \KL{q_i}{p_i}$ where $\KL{q_i}{p_i}$ is%
\footnote{See \cref{app:explicitboundexpressions} for the explicit expression of KL.
}
$
\klbin{\lambda_i}{\priorpruneprob_i} 
+ \frac {\lambda_i} 2
       \big( \psi(\sigma_i^f/\sigma_i^0) + {(W_i^f - W_i^0)^2}/{(\sigma_i^f)^2} \big ),
$
for $\psi(r) = r - 1 - \ln r$.
The first term penalizes $\sigma_i^f$ in terms of its multiplicative deviation from $\sigma_i^0$.
The second term penalizes the squared difference between the weights, scaled by the prior variance in that dimension.



Up until this point, 
we have not discussed loss functions. If the task is classification, the natural loss function is the zero--one loss function 
$\ell_{0-1}(\hat y,y) = 1_{\hat y \neq y}$.
However, since we will optimize the PAC-Bayes risk bound directly using gradient-based optimization,
we use cross entropy, a differentiable surrogate loss. 
When reporting bounds in figures, we must use zero--one loss in order to obtain valid bounds on classification risk.
Note that this procedure is valid since the bound is valid for any $Q$, even the one optimized on a loss different then the loss appearing in the bound.
See \cref{app:lossfunctions} for more details.

\section{PAC-Bayes Pruning}
\label{sec:stochastic_pruning}

We now describe \emph{PAC-Bayes Pruning (PBP)},
a general framework combining pruning, PAC Bayes, and data-dependent priors.
PAC-Bayes pruning proceeds in three stages:
\paragraph{Stage 1: Pre-train the prior.} Same as Stage 1 in PFT, but on a fraction $\DataPrior \subset S$ of the training data.
Obtain a deterministic predictor with weights $\hat{W}_P$. 
\paragraph{Stage 2: Fine-tune the prior (similar to PFT).}
Initialize the prior pruning probabilities $\hatpriorpruneprob_i$ based on one of the heuristic pruning methods described in \cref{sec:probpruning}. 
$\hatpriorpruneprob_i$ and  $\hat{W}_P$ correspond to our initialized values of $\priorpruneprob_i$ and $W_i^0$, respectively, in \cref{eq:prior}. 
From these initialized prior values, continue training by minimizing the empirical risk on $\DataPrior$ of the sparse stochastic network with distribution defined in \cref{eq:prior}, where minimization is performed over $W_0$,
$\{\priorpruneprob_i\}_{i=1}^D$, and $\{\sigma_i\}_{i=1}^{D}$. 

Unlike in PFT, the trained prior probabilities $\priorpruneprob_i$'s are recorded, and not thresholded to produce a deterministic pruning mask.

\paragraph{Stage 3: Optimize the PAC-Bayes bound in terms of posterior.}

Continue training the posterior $Q$, using all the training data $S$ to minimize the PAC-Bayes bound, \cref{eq:datadeppacbayesbound}, with respect to $\{\lambda_i\}_{i=1}^D,$ and $W^{f}_i$.

\begin{figure}
    \centering
    \includegraphics[width=0.5\linewidth]{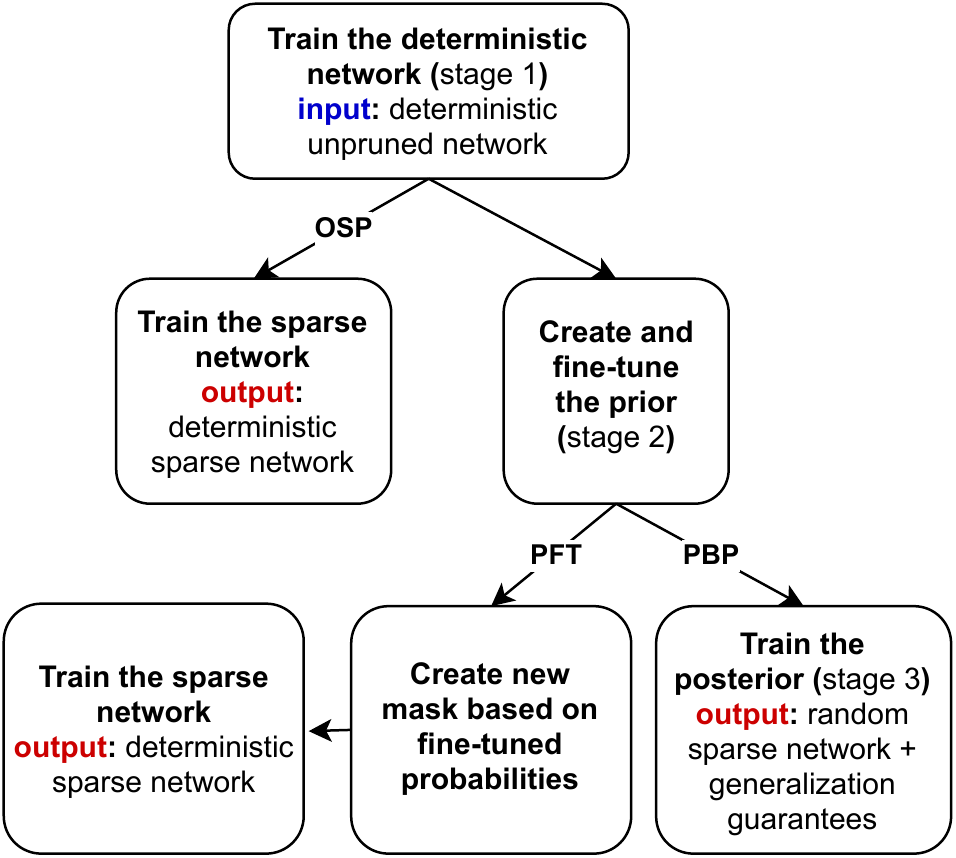}
    \caption{\small{OSP (One Shot Pruning), PFT (Probabilistic Fine-Tuning), and PBP (PAC-Bayes Pruning).}}
    \label{fig:pft_vs_pbp}
\end{figure}
It is worth noting that the above three-stage PBP algorithm encompasses the scope of our considerations in this work in full generality, 
and changing ``hyperparameters'' such as $|S_P|$ can and will lead to different procedures. 
For example, if we let $|S_P|{=}|S|$ and thus forgo stage 3, we see that stage 2 is by itself a viable method for improving a deterministic pruning mask (this is the essence of PFT), albeit without the generalization guarantees that the PAC-Bayes machinery offers. For example, one application of PFT could be to take a pruning mask obtained at initialization, and to fine-tune the mask, in addition to the weights, on training data. In this sense, PFT serves a different purpose to the \emph{self-bounded} PBP algorithm.\footnote{A self-bounded learning algorithm is an algorithm that outputs a predictor as well as a bound on the generalization error of that predictor \citep{freund1998self}.} 
\cref{fig:pft_vs_pbp} details different stages of PBP as a flowchart, highlighting the differences between PBP and PFT, as well as standard One-Shot Pruning (OSP).

To the best of our knowledge, PFT is the first algorithm to combine deterministic pruning with probabilistic fine-tuning in order to find better local masks that boost accuracy and PBP is the first algorithm to combine stochastic pruning and PAC-Bayes optimization. One benefit offered by PBP's combination of PAC-Bayes and pruning is access to tight generalization guarantees that can outperform held-out validation bounds by using all available training data in the training process and bound (\cref{sec:main_exps}).

\subsection{Bound optimization as constrained feature realignment}
We now return to our linear model analysis in light of Stage (3) of the PBP algorithm.
We highlight a few key changes, but refer the reader to \cref{sec:optpacbayesfeature} for all details.
First, PAC-Bayes risk bound has an additional regularizer due to the KL term. The effect of this term is to force $\lambda_i$ to stay close to its prior value $\priorpruneprob_i$. Second, the growth of $\lambda_i$ depends on the agreement between the data $S_P$ and $\SbarP$: 
if the sign of the gradient on $\SbarP$ disagrees with the sign of the weight (learned from $S_P$), then $\lambda_i$ is driven to zero. 
Otherwise, $\lambda_i$ increases.
In \cref{sec:optpacbayesfeature}, we capture this agreement between
$S_P$ and $\SbarP$ via a measure of data distribution complexity/stability.
In \cref{lemma:2} (\cref{app:kltermondrift}), we provide conditions under which the bound grows linearly or quadratically with this measure of complexity (See \cref{app:kltermondrift}).

Note that our analysis of linear models under squared error extends to neural networks in the Neural Tangent Kernel regime (i.e., the infinite-width limit). See \cref{app:NTK} for more details.

\section{Experiments: PAC-Bayes Pruning}
\label{sec:pbpexperiments}

Recall that full details on the networks, datasets,  hyperparameters, optimizers, and loss functions are provided in \cref{app:experimentaldetails}. 
\cref{app:additionalresults} contains a number of additional experiments on PAC-Bayes Pruning (PBP), including the trade-off between the tightness of the bound and the test error for PBP, the trade-off between overall mask entropy and performance in PBP,
an analysis of the evolution of the prior and posterior probabilities across the different stages of PBP, and an analysis of the deterministic masks created from One Shot Pruning methods.

\subsection{Self-bounded learning via PAC-Bayes} \label{sec:main_exps}
\begin{figure}[t]
    \centering
    \includegraphics[width=1.\linewidth]{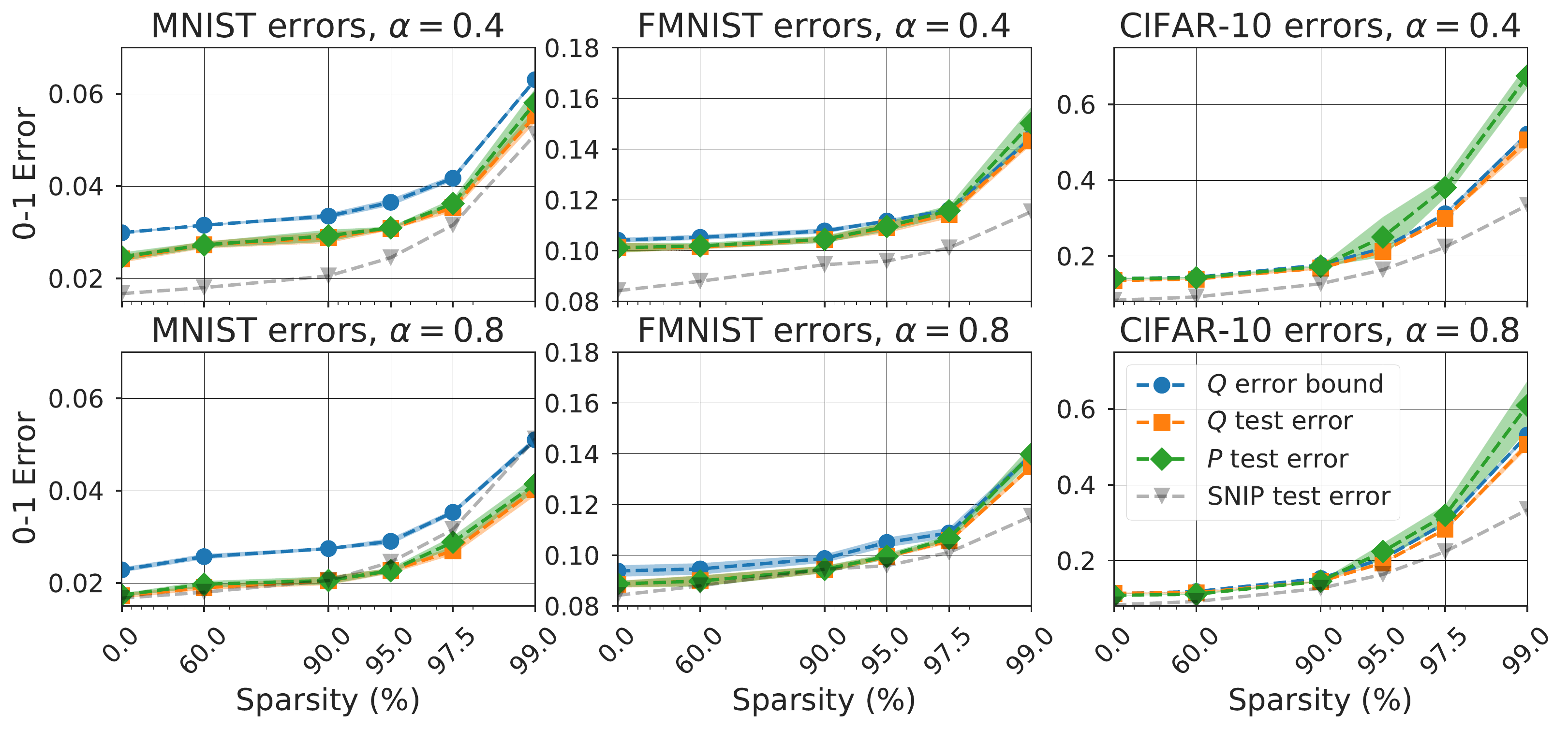}
    \caption{\small{Error and error bound versus sparsity on image classification tasks for different $\alpha=|S_P|/|S|$. Displayed intervals correspond to 95\% confidence over 5 independent seeds.} }
    \label{fig:EvS}
\end{figure}

We evaluate our proposed stochastic pruning with PAC Bayes framework (PBP), which returns a stochastic predictor in contrast to many other pruning approaches. We consider MNIST, Fashion-MNIST \& CIFAR-10\footnote{Datasets links:   \href{http://yann.lecun.com/exdb/mnist/}{MNIST}, \href{https://github.com/zalandoresearch/fashion-mnist}{FMNIST},
\href{https://www.cs.toronto.edu/~kriz/cifar.html}{CIFAR-10}.} 
with corresponding architectures to \cref{sec:PFT}. In \cref{fig:EvS}, we compare the PAC-Bayes generalization bounds on 0-1 error (blue) of our stochastic predictor $Q$ to the corresponding observed test set error (orange) and the observed test error of our sparse stochastic prior P (green) over a range of sparsities and pre-training dataset fraction $\alpha$. 
For reference, we also plot the test error of a baseline pruning method, SNIP \citep{lee2018snip}, to compare our test errors and bounds to a standard \emph{deterministic} OSP method which doesn't provide generalization guarantees.  
In general, we see that a higher $\alpha$ value not only improves the bound, but also decreases error of randomized predictors $P$ and $Q$. 
More data-dependence helps across different sparsity levels and datasets.

We also observe that the bounds on the error of $Q$, and the error of $Q$ as measured on the test set are within a few percentage points of our baseline for a large range of sparsity levels.
The gap between the baseline and the performance of $Q$ increases with sparsity for both values of $\alpha$.
Interestingly, when trained on CIFAR-10, we also see that the posterior $Q$ significantly outperforms the prior $P$, obtaining 10\% lower error when $\alpha=0.4$, and 97.5\% of the weights are pruned. 
In this case, since $Q$ test error bound is already lower than the prior test error, any held out dataset bound on the performance of $P$ would be worse than our PAC-Bayes bound on $Q$. This demonstrates the strength of our proposed PBP algorithm in that it uses all of the available data for training in order to provide generalization bounds. This is crucial in regimes of low or complex data where we expect to need all of the available data for training a good classifier.

\section{Related Work}
Recent work by \citet{elesedy2020lottery} has shown that, in the context of linear models, magnitude pruning zeros out the weights based on the magnitude of feature alignment under certain assumptions on the feature covariance matrix. Our analysis is complementary to these findings. 

\paragraph{Other pruning methods.}
Our work uses a class of pruning
approaches that identify the set of weights to prune based on a criterion applied to each weight by the end of a single training run. We refer to such methods as one-shot pruning (OSP). These include magnitude pruning \citep{han2015learning} and SNIP \citep{lee2018snip}, as well as many others (e.g., see GraSP \citep{wang2020picking} and SynFlow \citep{tanaka2020pruning} for pruning at initialization, and $L_0$ regularization-based  \citep{savarese2019winning,gale2019state,louizos2017learning} and dropout-based \citep{molchanov2017variational,zhu2017prune} for pruning during training).
SNIP was proposed as a technique to prune at initialization, with the remaining weights trained as usual. In contrast, magnitude pruning was proposed as a technique to apply after training, following by fine-tuning. 
However, state-of-the-art applications of magnitude pruning now employ iterative magnitude pruning (IMP) \citep{frankle2018lottery,frankle2020linear}, where a fraction of the smallest weights are pruned on each round, followed by rewinding training back to an earlier iteration \citep{frankle2020linear} or simply rewinding the learning rates \citep{renda2020comparing}. \citet{gale2019state,frankle2020missing} show that none of the OSP techniques dominate one another, and IMP outperforms them all \citep{renda2020comparing}.

The pruning method that is most related to PFT is an extension of variational dropout due to \citet{molchanov2017variational}.
In variational dropout,  the weights are multiplied by Gaussian variables instead of Bernoulli variables. Although the variances of the Gaussian mask can be forced to be small or large, the corresponding model is never sparse (with probability one), and another thresholding step is required to make the model sparse (e.g., remove the weight with the highest Gaussian mask variance). Another major difference is the choice of the prior; Gaussian dropout consider the improper log-scale uniform prior since it is the only prior that makes variational inference consistent with Gaussian dropout. As a result, this prior can only yield vacuous PAC-Bayes bounds.
\section{Discussion and Limitations}\label{sec:limitations}

In this work, we introduced a new algorithmic framework for stochastic pruning. The generality of this framework allowed us to derive two algorithms, PFT and PBP, that act on the same input network and return different outputs. We elucidate the non-optimality of OSP pruning methods by showing that probabilistic fine-tuning (PFT) finds better local masks, starting from popular OSP methods. To depict this gain in performance, we analyze PFT in the context of linear models and show that PFT learns pruning probabilities based on feature alignment. Further, we show a regularization effect on pruning probabilities arising from minimizing risk of a stochastic predictor, rather than an average one. This is complementary to \citep{WagerWangLiang2013}, where the authors demonstrate a similar regularization effect on the weights when training a stochastic predictor. On small vision tasks, we notice that PFT consistently boosts the accuracy of existing One-Shot Pruning methods (\cref{sec:pftexperiments}), however, the computational cost of PFT is no less than ordinary training and so scales with network size. Adapting PFT for large scale tasks is an interesting topic for future work.

Motivated by the apparent robustness of pruned networks to perturbations of the weights, we introduced PAC-Bayes Pruning (PBP), which is a self-bounded learning algorithm with data-dependent priors, aimed at training a sparse \emph{stochastic} network. In the linear model context, we demonstrate that the tightness of a data-dependent PAC-Bayes bound obtained via PBP depends on how the feature alignment changes as we add more data to the learning problem.
In our experiments, we obtained tight numerical bounds for sparse DNNs on a variety of datasets as demonstrated in \cref{sec:pbpexperiments}. However, we observe that stochastic predictors seem to consistently under-perform their deterministic counterparts. The role of product prior distributions on the weights and pruning probabilities is worth investigating. The consistent drop in performance seen when pruning weights highlights the importance of considering the interaction between weights when pruning. 

\section*{Acknowledgements} 
The authors would like to thank Blair Bilodeau, Bryn Elesedy, Jonathan Frankle, and Daniel M. Roy for feedback on drafts of this work.


\printbibliography


\appendix



\section{Additional background material}
\subsection{Initializing pruning probabilities}\label{app:initializing_pruning_probs}
Below we discuss some choices of initializing pruning probabilities $\priorpruneprob$. The main difference between these choices is whether $\priorpruneprob$'s are chosen based on some criterion $g$, which we refer to as \emph{probabilistic pruning based on the criterion $g$}, or randomly.

Let $s$ be the sparsity that we want to achieve in the probabilistic model, i.e., 
the average of the entries of $1 - \priorpruneprob$.
We consider two versions of setting $\priorpruneprob$:

\begin{itemize}
    \item \emph{Isotropic $P$ (random pruning)}: 
    the weights have equal probability to be pruned. To achieve an average sparsity $s$, we set $\priorpruneprob_i = 1-s$.
    \item \emph{Block isotropic $P$}: 

    we divide the weights into two blocks based on their $g_i$ value (evaluated on a deterministic network $W_0$).
 
    Letting $k = (1-s)D$, $g^{(k)}$ be the $k^{th}$ order statistic of the sequence $(g_i)_{1\leq i \leq p}$, $\epsilon \in (0,1)$, and $s\epsilon/(1-s)<1$, we choose 
    \begin{equation}\label{eq:blockisotropicprior}
    \priorpruneprob_i =
    \begin{cases*}
      1 - s\epsilon/(1-s) & if $g_i \geq g^{(k)}$ \\
      \epsilon        & otherwise.
    \end{cases*}
    \end{equation}
    When $\epsilon \ll 1$, this prior allows small fluctuations around the deterministic pruned network.
\end{itemize}
Both of these choices induce $P$ with an expected sparsity $s$. 

\subsection{A standard data-independent PAC-Bayes bound}

Here we introduce a classical PAC-Bayes bound and connect it to the one used in \cref{thm:pac_bayes_data_dependent}.

Writing $dQ/dP$ for the Radon--Nikodym derivative of $Q$ with respect to $P$, recall that the 
Kullback-Leibler divergence (or entropy)
of $Q$ (relative) to $P$, is defined to be
$
\KL{Q}{P} = \int\log ({dQ}/{dP}) dQ
$
when $Q$ is absolutely continuous with respect to $P$ 
and infinite otherwise.
For $p, q \in (0,1)$, let $\klbin{q}{p}$ denote the KL divergence of the Bernoulli distribution $\mathcal{B}(q)$ to $\mathcal{B}(p)$.
Thus, $\klbin{q}{p} = q \log({q}/{p}) + \bar q \log(\bar q/\bar p)$, where $\bar p = 1-p$ and $\bar q = 1-q$.

The following PAC-Bayes theorem, 
of \citet{langford2001pacbayes},
is stated in terms of the inverse Bernoulli KL divergence,
$$
\klinv{a}{\varepsilon} = \sup\{ p \in [0,1] : \text{kl}(a||p) \leq \varepsilon \}\,.
$$
\begin{thm}
\label{thm:pac_bayes}
For every $[0, 1]$-valued loss function $\ell$, 
data distribution $\Dist$, natural $N \in \mathbb{N}$, distribution $P$ on weights, and probability $\delta \in (0, 1)$,
with probability at least $1-\delta$ over training data $S \sim \Dist^N$,
\begin{equation}
    \forall Q,\, \Risk{\Dist}{Q} \leq 
    \klinv{  \EmpRisk{S}{Q} }{ \smash{ \frac{\KL{Q}{P} + \log(\frac{2 \sqrt{N}}{\delta} ) }{N}} }.
    \label{eq:standardpacbayesbound}
\end{equation}
\end{thm}

Substituting our concrete KL term into \cref{thm:pac_bayes}, we obtain a quantity that we can explicitly optimize in terms of the parameters that define the posterior. Note that gradients of the empirical Gibbs risk term are in general intractable. However, as is standard, tractable stochastic gradients can be obtained. We describe our approach in \cref{app:gumbel_softmax}.

Note that the inverse KL divergence in \cref{thm:pac_bayes} does not admit an analytical form. 
Different relaxations exist in the literature.
In our experimental setting, we optimize the bound stated in \cref{thm:pac_bayes_data_dependent}.

\subsection{Surrogate loss function}
\label{app:lossfunctions}

Though our bounds are evaluated on the 0-1 error, this is not a differentiable loss function. 
We follow \citep{dziugaite2020role} in using gradient-based optimization with the average cross entropy loss as a surrogate empirical risk $\EmpRisk{S}{Q}$ during posterior training. 
We clamp the probabilities smaller than $10^{-4}$ to ensure that the optimization objective remains bounded.

We would like to reiterate that the following produces a valid procedure since \cref{thm:pac_bayes_data_dependent} is valid for \emph{any} posterior $Q$, trained in any way: we can compute $Q$ via optimization of \cref{eq:datadeppacbayesbound} with a cross entropy loss, and evaluate the bound with the performance of $Q$ on 0--1 loss.

Also note that the evaluation of a PAC-Bayes bound involves the empirical risk $\EmpRisk{S \setminus S_P}{Q}$. In general, the computation of this quantity is intractable. However, a confidence interval can be obtained using a Chernoff bound.

\section{Algorithmic details}

\subsection{Posterior probability mapping.}
Optimizing the posterior probabilities $\lambda$ might cause numerical issues whenever the gradient update pushes the probability out of the interval $(0,1)$. 
To avoid such scenarios, we use a parameterization of the posterior probabilities that ensures probabilities remain in $(0,1)$. We consider two options
\begin{itemize}
\item The Sigmoid function: $\lambda_i = \textrm{Sig}(\lambda_i')$
\item The Clamp function:
$\lambda_i = \begin{cases}
\lambda_i \quad \lambda_i \in (0,1)\\
1 \quad \lambda_i \geq 1 \\
0 \quad \lambda_i \leq 0
\end{cases} $
\end{itemize}

The sigmoid function tends to outperform the clamp function when the prior is chosen to be a small perturbation of the deterministic sparse network, e.g., block isotropic prior with small $\varepsilon$. However, the clamp function performs better than sigmoid when the prior is sufficiently random (i.e., a significant fraction of the weights have pruning probabilities not close to 0 or 1). The results reported in \cref{fig:EvS} use Sigmoid mapping while those reported in \cref{fig:pft_results} use the clamp function. These choices were found to be optimal by hyperparameter search.

\subsection{Optimization Algorithm}
\label{app:optalg}

The optimization of the empirical risk requires computing the gradients w.r.t the posterior mean weights $W^f$ and the posterior pruning probabilities $\lambda$. The direct computation of the latter is problematic from a practical perspective. To see this, recall that the empirical risk is given by 
$$
\EmpRisk{S}{Q} = \EE_{b, W}[\SurEmpRisk{S}{b \circ W}],
$$
where $W$ and $b$ are the weights and the masks, and $\SurEmpRisk{S}{b \circ W} = \frac{1}{N} \sum_{i=1}^N \surloss(f_{b \circ W}(x), y)$ is the empirical risk for a surrogate differentiable loss $\surloss$.

For some $i$, let $e_i(b'_i) = \EE_Q[\SurEmpRisk{S}{b\circ W} |b_i = b'_i]$. Then, we have that
$$
\SurEmpRisk{S}{Q} =  \lambda_i e_i(1) + (1-\lambda_i) e_i(0)
$$
this yields
$$
\partial_{\lambda_i} \SurEmpRisk{S}{Q} =  e_i(1) - e_i(0)
$$
Computing the gradient with this formula is impractical as it requires an extensive number of forward passes ($2\times p\times k$ where $k$ is the number of network samples used to estimate $e_i(1)$ and $e_i(0)$). The conditional expectation makes it impractical to obtain a direct Monte Carlo estimate of $\partial_{\lambda_i} \SurEmpRisk{S}{Q}$; indeed, we have to fix $b_i = 1$ or $b_i = 0$ and simulate the rest of the mask, in order to estimate $\partial_{\lambda_i} \SurEmpRisk{S}{Q}$. Repeating this procedure for all $i$'s would require extensive computational resources. However, we can also simulate $k$ mask samples and use these samples to estimate $\partial_{\lambda_i} \SurEmpRisk{S}{Q}$, i.e., find the sample masks where $b_i = 1$ and use them to estimate $e_i(1)$, and vice-versa for $e_i(0)$. Algorithm 1 illustrates this procedure. Unfortunately, this algorithm would still require tremendous number of samples to update extreme probabilities, which limits its benefits.

Fortunately, a relaxation of this optimization problem known as the Gumbel--Softmax trick makes the gradient easier to approximate using standard Monte Carlo estimate. We describe this relaxation in details in the next section.

\begin{algorithm}[t]
\SetAlgoLined
\KwResult{Approximate optimization of PAC-Bayes bound $E$}\label{alg:alg1}
 prior $P$, learning rate $\eta$, sample size $m$\;
 \While{not converged}{
    \For{k=1 to m}{
        simulate $w^k_i \sim \NormalD{W^f_i}{s_i^2}$ for all $i$\;
        simulate Bernoulli variables $B_k \sim \mathcal{B}(\lambda_i)$ for all $i$\;}
    $a_i^k = \frac{B^k_i}{\sum_k B^k_i}$\;
    $b_i^k = \frac{1 - B^k_i}{\sum_k (1 - B^k_i)}$\;
    
    $ \zeta_i =\frac{1}{m} \sum_k \left[ \mathcal{L}((w^k_j B^k_j)_{j\neq i}, w^k_i) - \mathcal{L}((w^k_j B^k_j)_{j\neq i}, 0) \right] \approx \sum_{k=1}^m (a_i^k + b_i^k)(2 B^k_i -1)  \mathcal{L}((w^k_j B^k_j)_{1\leq j\leq p})$\;
    $ \beta_i = \frac{1}{m} \sum_k \partial_{w^k_i}\mathcal{L}((w^k_j B^k_j)_{1\leq j\leq p})$\;
    $\lambda_i \longleftarrow \lambda_i  - \eta (\zeta_i + \partial_{\lambda_i} \Gamma(Q))\times \mathrm{1}_{0<\sum_{k} B^k_i < m}$\;
    $W^f_i \longleftarrow W^f_i - \eta ( \beta_i +   \partial_{W^f_i} \Gamma(Q))$\
 }
 \caption{Optimization of PAC-Bayes bound}
\end{algorithm}
\subsection{Reparameterization trick and Gumbel-Softmax}\label{app:gumbel_softmax}
The empirical risk $\SurEmpRisk{S}{Q}$ depends on $\lambda$ in an implicit way, thus, it is not straightforward to compute the gradient  $\partial_{\lambda} \SurEmpRisk{S}{Q}$ using Monte Carlo samples. 
A classic approach is the \emph{reparameterization trick} which consists of finding a mapping that makes the dependence on $\lambda$ explicit. 
In the case of a discrete distribution (Bernoulli in our case), a common reparameterization trick is Gumbel--Max (GM) trick \citep{Jang2017,Madd2017}, which can be traced back to \citet{luce1959gumbel}. 
It was introduced as a method for sampling from discrete random variables using explicit dependence on the probabilities of each state. 
More precisely, consider a discrete representation $d \in \{0,1\}^n$ such that $\sum_{i=1}^n d_i = 1$ and let $(\priorpruneprob_1, .., \priorpruneprob_n) \in (\NNReals^{*})^n$ denote an unnormalized probability vector. 
The GM trick can then be described by the following steps:
\begin{enumerate}
    \item sample $U_k \sim \mathcal{U}(0,1)$ \iid for $k\in [n]$;
    \item find $k = \arg \max_{k} \{\log(\priorpruneprob_k) - \log(- \log(U_k)) \}$, set $D_k=1$ and $D_i=0$ for $i\neq k$.
\end{enumerate}
Then, we have $$\mathbb{P}(D_k = 1) =\frac{\priorpruneprob_k }{\sum_{i=1}^n \priorpruneprob_i}.$$
The name Gumbel--Max trick was inspired by the fact that the random variable $-\log(-\log(U))$ has a Gumbel distribution.

Although the GM trick allows straightforward simulation of discrete variables, it is not practical for gradient computing since the the differentiation of the $\max$ function might cause numerical issues. For this purpose, \cite{maddison2017concrete} introduced the Gumbel--Softmax (GS) trick which relaxes the discrete distribution to the \emph{concrete} distribution. It proceeds as follows:  
\begin{itemize}
    \item fix $\beta \in \NNReals^*$ and sample $G_k \sim \textrm{Gumbel}$ \iid;
    \item set $X_k = \frac{\exp{((\log(\priorpruneprob_k) + G_k)/ \beta)}}{\sum_{i=1}^n\exp{((\log(\priorpruneprob_i) + G_i)/ \beta)}} .$
\end{itemize}
Then, $X$ has a concrete distribution which we denote $Concrete(\priorpruneprob_1, ..., \priorpruneprob_n)$. 
Further, it follows that $$ \lim_{\beta \to 0} \mathbb{P}(X_k > X_i \textrm{ for all } i\neq k) = \frac{\priorpruneprob_k}{\sum_{i=1}^n \priorpruneprob_i}.$$ 

\paragraph{The relaxed problem.} In our case, given a sequence of independent random variables $X_i \sim Concrete(\lambda_i, 1-\lambda_i)$, 
we optimize the relaxed empirical risk, given by 
$\EmpRisk{S}{(X_i, W_i)_{1\leq i \leq p}}.$ We take $\beta=0.5$ throughout this work.

\subsection{Kullback–Leibler divergence for spike and slab distributions}
\label{app:explicitboundexpressions}

With the spike and slab prior $P$ and posterior $Q$, the KL divergence is tractable and is given by
\begin{equation*}
\begin{split}
\KL{Q}{P} &= \sum_{i=1}^D KL(q_i||p_i) \\
&= \sum_{i=1}^D \klbin{\lambda_i}{\priorpruneprob_i} + \lambda_i \underbrace{ \KL{\NormalD{W^f_i}{s_i^2} }{ \NormalD{W^0_i}{\sigma_i^2} }}_{\gamma_i},
\end{split}
\end{equation*}
where for all $1\leq i\leq D$,
\begin{align*}
\klbin{\lambda_i}{\priorpruneprob_i} &= (1-\lambda_i) \log\left(\frac{1-\lambda_i}{1-\priorpruneprob_i}\right) + \lambda_i \log\left(\frac{\lambda_i}{\priorpruneprob_i}\right),\label{eq:klbinarylambda}\\
\gamma_i &= \frac{1}{2}\left( \frac{(W_i^f - W_i^0)^2}{\sigma_i^2} + \frac{s_i^2}{\sigma_i^2} - \log(\frac{s_i^2}{\sigma_i^2}) -1 \right).
\end{align*}

\section{Linear Model Analysis}
\label{app:linmodel}

\subsection{Learning prior pruning probabilities}\label{app:learning_prior_probs}

Here we provide the proof of \cref{lemma1:statement}.

\begin{proof} Consider a related empirical risk of an average predictor, in this case, a linear model with weights $\mweights$. 
Then \cref{eq:derivativeofemprisklinear} can also be expressed in terms of the gradient of $\EmpRisk{S_P}{\mweights}$ with respect to $\mweights_i$. 
In particular,
\begin{equation}
    \frac{\partial \EmpRisk{S_P}{P}}{\partial \priorpruneprob_i} 
    = \weight_i \frac{\partial\EmpRisk{S_P}{\mweights}}
                     {\partial \mweights_i} 
    + r_i^0,
    \label{eq:minriskgibbs}
\end{equation}
where $r_i^0 = \frac 1 2 (\weight_i)^2 (\Sigma_{P})_{ii}$.
Now consider minimizing the empirical risk of the average classifier, $\EmpRisk{S}{\mweights}$,  with respect to $\priorpruneprob_i$. By the chain rule, the gradient is
$
    \frac{\partial \EmpRisk{S}{\mweights}}{\partial \priorpruneprob_i} 
    = \weight_i \frac{\partial \EmpRisk{S}{\mweights}}{\partial \mweights_i}.$
It follows that
 $   \frac{\partial \EmpRisk{S_P}{P}}{\partial \priorpruneprob_i}  - \frac{\partial \EmpRisk{S_P}{\mweights}}{\partial \priorpruneprob_i}  = r_i^0.$
\end{proof}

\subsubsection{Pruning probabilities are aligned with weight magnitude growth}
\label{subsec:weightmag}

\emph{Here we show that early stopping induces a pruning criterion that is aligned with the magnitude of the weights. In linear regression with diagonal feature covariance, the induced pruning criterion is based on variance-normalized feature alignment.}

From \cref{eq:minriskgibbs}, we can conclude the following: the first gradient term increases $\lambda_i$ when
the gradient of the empirical risk of $\mweights$ pushes the weight magnitude up, and decreases otherwise. 
The regularization term drives $\lambda_i$ to zero. 

Note that the weights minimizing the empirical risk $\EmpRisk{S}{w}$ under square loss are 
\begin{equation}
    w_i = \frac{1}{|S|} \Sigma^{-1}(X) \featurei{X}{i}^T Y, 
    \label{eq:gradflowweights}
\end{equation}

In linear regression, the weight gradient depends on the feature alignment -- large feature alignment magnitude yields large magnitude weights (up to rescaling based on $\Sigma_P$), as seen from \cref{eq:gradflowweights}. 
When $\Sigma_P$ is diagonal, 
at convergence 
$w_i = (\Sigma_P)_{ii}^{-1} A_i(P)$ and $\frac{\partial \EmpRisk{S_P}{P}}{\partial \priorpruneprob_i} = \frac{1}{2} (w_i^P)^2$. 
If we start by initializing the pruning probabilities all to the same value, 
then $\priorpruneprob_i$'s with larger (always non-negative) gradient increase more rapidly, and thus can be interpreted as being more important. 
With early stopping, this yields a pruning criterion based on the magnitude of feature alignment, normalized by $(\Sigma_P)^2_{ii}$. 
In other words, the weights that have the highest feature alignment (after adjusting for the variance), are the ones that are not pruned. Magnitude-based pruning and feature alignment connections were recently described by \citet{elesedy2020lottery}.

\paragraph{Non-diagonal covariance.}
We extend the analysis when the features are correlated. Assume $(\Sigma_P)_{ij} = (\Sigma_P)_{ji} \neq 0$ for some $i \neq j$ and assume $(\Sigma_P)_{i'j'} = 0$ for all other $i' \neq i,j', \, j' \neq j$. We now compute the ratio of ${\priorpruneprob_i}/{\priorpruneprob_j}$, indicating the relative importance of $w_i$ and $w_j$. 

Setting the gradient $\frac{\partial \EmpRisk{S_P}{P}}{\partial \priorpruneprob_i}$ to 0, if $w_i \neq 0$, we have
\begin{equation}
    \priorpruneprob_j = \frac{2A_i(P) - w_i (\Sigma_P)_{ii} }{2 w_j (\Sigma_P)_{ij}}.
    \label{eq:optimalpriorpruneprob}
\end{equation}

Using  \cref{eq:gradflowweights}, we can also write $w_i$ for this special $\Sigma_P$ case as 
\begin{equation}
    w_i = (\Sigma_P)_{ii} A_i(P) - (\Sigma_P)_{ij} A_j(P).
    \label{eq:wicomputedexplicitely}
\end{equation}
Then for $(\Sigma_P)_{ii} = (\Sigma_P)_{jj} = 1$,  by substituting \cref{eq:wicomputedexplicitely} to \cref{eq:optimalpriorpruneprob}, we can express the ratio as
\begin{equation}
\frac{\priorpruneprob_i}{\priorpruneprob_j} = \frac{(c_{ji} + (\Sigma_P)_{ij} ) (1/c_{ji} - (\Sigma_P)_{ij} ) }{(c_{ji} - (\Sigma_P)_{ij} ) (1/c_{ji} + (\Sigma_P)_{ij} ) },
\label{eq:ratiooflambdas}
\end{equation}
where $c_{ji} = \frac{A_j(P)}{A_i(P)}$.\footnote{Note, that in \cref{eq:ratiooflambdas}, to obtain valid values, we need to take into account that the magnitudes of $(\Sigma_P)_{ij} $ and $c_{ji}$ are dependent.}
Studying \cref{eq:ratiooflambdas}, we conclude the following:
\emph{
\begin{itemize}
    \item if $\sign(A_i(P)) \cdot \sign(A_j(P)) = \sign(\Sigma_P)_{ij} $, then the weight with a smaller magnitude of feature alignment is more likely to be pruned;
    \item otherwise, the weight with larger feature alignment is more likely to be pruned. 
\end{itemize}}

Intuitively, if the correlation $(\Sigma_P)_{ij}$ is positive, but feature alignment signs $\sign(A_i(P)) \neq \sign(A_i(P))$, then the weight $w_k$ with a higher magnitude of feature alignment $|A_k(P)|$ is likely to be misleading and we should assign a smaller probability $\priorpruneprob_k$ (for $k \in \{i,j\}$).

\subsection{Optimizing the PAC-Bayes bound as constrained Feature Realignment}
\label{sec:optpacbayesfeature}

In this section, we analyze a three-stage process, the first two steps being probabilistic fine-tuning:

\begin{enumerate}
    \item Train the linear model with $(X_{P}, Y_{P})$. 
    Let $\weight$ denote the trained weights.
    \item Learn pruning probabilities $\{ \priorpruneprob_i\}_{i=1}^D$ by minimizing $\EmpRisk{S}{P}$, where $P$ is a product distribution $\prod_{i=1}^D p_i$, with $p_i = (1-\priorpruneprob_i) \delta_{\{0\}} + \priorpruneprob_i \delta_{\{\weight_i\}}$.

    \item Optimize a PAC-Bayes bound with prior P over the $D$ weights with respect to a product posterior $Q$, where the posterior distribution on each weight $i \in [D]$ is of the form
        $q_i = (1-\lambda_i) \delta_{\{0\}} + \lambda_i \delta_{\{\weight_i\}}$.
        (That is, we optimize over the $\lambda_i$'s.)

\end{enumerate}

    To ease our analysis, we consider a linear PAC-Bayes bound due to \citep{catoni2007pacbayes}.
    The resulting optimization objective for our choice of prior and posterior is
    \begin{equation}
    B(\lambda) = \EmpRisk{S}{Q} + \frac{\zeta}{n} \sum_{i=1}^D \KL{q_i}{p_i},
    \label{eq:linobjective}
    \end{equation}
    where $\KL{q_i}{p_i} = \klbin{\lambda_i}{\priorpruneprob_i}$ (see \cref{app:explicitboundexpressions}).

Using \cref{eq:minriskgibbs}, we express
\begin{equation}
    \frac{\partial B}{\partial \lambda_i} 
    = \weight_i \frac{\partial\EmpRisk{\SbarP}{\mweights}}
                     {\partial \mweights_i} 
    + r^0_i
    + r^{\priorpruneprob}_i,
    \label{eq:pacbayesboundderivative}
\end{equation}
where $r^{\priorpruneprob}_i = \frac{\partial \klbin{\lambda_i}{\priorpruneprob_i}}{\partial \lambda_i}$ is the KL penalty. 
The penalty term acts as a regularizer, forcing $\lambda_i$ to stay close to its prior value $\priorpruneprob_i$. 
\cref{fig:kl_penalty} illustrates this effect: when $\lambda_i > \priorpruneprob_i$, the penalty term is positive, driving the value of $\lambda_i$ to be lower, and vice versa. This is in addition to the regularization coming from $r^0_i$, which drives $\lambda_i$ to zero.

In contrast to the analysis in \cref{sec:learning_prior_probs}, here the gradient of the empirical risk of $\mweights$ is computed on the new data $\SbarP$. If the sign of the gradient on $\SbarP$ disagrees with the sign of the weight (that was optimized on $S_P$),  $\lambda_i$ will be driven to zero. Otherwise, $\lambda_i$ will increase. 

When $\Sigma(X_{\Bar{P}}) = \mathbb{I}$, this translates to the sign of the feature alignment $A_i(\bar P)$ matching the sign of $\weight_i$. 
Further, if $\weight_i$ is the minimizer of the empirical risk of a deterministic predictor, then $w_i = A_i(P)$. 
In this case, $\lambda_i$ is driven to zero if $\sign(A_i(P)) \neq \sign(A_i(\bar P))$, and to 1, otherwise.

\subsubsection{Closed-form solution for uncorrelated features}

\begin{figure}[tbh]
    \centering
    \includegraphics[width=0.5\linewidth]{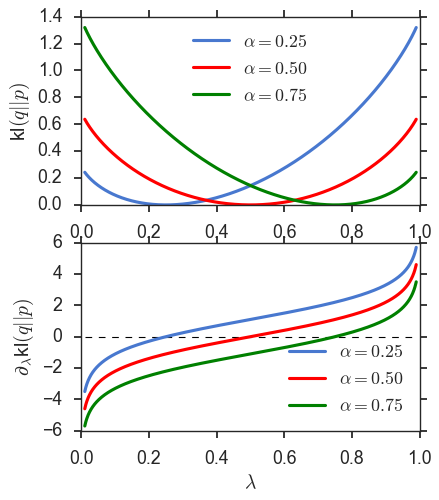}
    \caption{KL penalty for Bernoulli random variables.}
    \label{fig:kl_penalty}
\end{figure}

Assume $\Sigma(X_{\Bar{P}}) = \Sigma(X_{P}) = \mathbb{I}$ and let $\kappa = \frac{\zeta}{N-M}$. 
The gradient of $\lambda_i$'s when minimizing \cref{eq:linobjective} is given by 
\begin{equation}\label{equation:gradient_flow}
\frac{d \lambda_{i,t}}{ dt}= -\mu \left( \Drift + \kappa \,  \frac{\partial \KL{q_i}{p_i}}{\partial \lambda_i}\right),
\end{equation}

where $\mu>0$ is the learning rate 
\\
The KL divergence above is reduced to that of a Bernoulli variable, and for each $i \in [D]$ is given by
$$
\klbin{\lambda_i}{\priorpruneprob_i} = \lambda_i \log\left(\frac{\lambda_i}{\priorpruneprob_i}\right) + (1-\lambda_i) \log\left(\frac{1-\lambda_i}{1-\priorpruneprob_i}\right),
$$
with derivative 
$$
\frac{\partial \klbin{\lambda_i}{\priorpruneprob_i}}{\partial \lambda_i} = \log\left(\frac{\lambda_i}{\priorpruneprob_i}\right) - \log\left(\frac{1 - \lambda_i}{1 - \priorpruneprob_i}\right).
$$
The penalty in the objective \cref{eq:linobjective} coming from this KL term and its derivative are plotted in \cref{fig:kl_penalty} as a function of $\lambda_i$'s, posterior pruning probabilities, for a few different values of $\priorpruneprob_i$.

The gradient flow of $\lambda_i$ minimizing the bound converges to the unique solution given by

\begin{equation}
\lambda_{i,\infty} = \frac{\priorpruneprob_i}{\priorpruneprob_i + (1-\priorpruneprob_i) \exp(\kappa^{-1} \Drift)},
\label{eq:lambdaoptimal}
\end{equation}
where $\Drift = r_i^0 - \weight_i A_i(\bar P)$, i.e., the first two terms in \cref{eq:pacbayesboundderivative}.

For $\Drift = 0$, we recover the value $\lambda_{i,\infty} = \priorpruneprob_i$.
\paragraph{Approximations.} For particular values of the drift, $\lambda_{i,\infty}$ can be approximated with Taylor expansion:
\begin{itemize}
    \item $|\Drift| \ll 1$: $\lambda_{i,\infty} \approx \priorpruneprob_i (1 + \kappa^{-1} \Drift (1-\priorpruneprob_i))$;
    \item  $\Drift \ll -1$: $\lambda_{i,\infty} \approx (1 - (\priorpruneprob_i)^{-1} (1 - \priorpruneprob_i) \exp(\kappa^{-1} \Drift))$;
    \item $\Drift \gg 1$: $\lambda_{i,\infty} \approx \priorpruneprob_i (1 - \priorpruneprob_i)^{-1} \exp(- \kappa^{-1} \Drift)$ .
\end{itemize}
In words, when the drift magnitude is large, the bias coming from the value of $\priorpruneprob_i$ learned on the data $X_P$ diminishes as $N-M$ or $|\Drift|$ grow:
$\lambda_{i,\infty}$ tends to 1 (large negative drift) or 0 (large positive drift). 


\subsubsection{Concentration of feature alignment as a measure of data distribution complexity}

The tightness of the PAC-Bayes generalization guarantee is determined by the size of $\kappa \sum_i \klbin{\lambda_i}{\priorpruneprob_i}$. 
Here we state the dependence of the KL term on the drift term.
\begin{lemma}
\label{lemma:2}
Under certain approximations, the tightness of the data-dependent PAC-Bayes bound on the generalization error scales with the drift term:
\begin{itemize}
    \item with $|\Drift| \ll 1$, the bound is quadratic in $\Drift$;
    \item with $|\Drift| \gg 1$, the bound is linear in $\Drift$.
\end{itemize}
\end{lemma}

For the details, see \cref{app:kltermondrift}.

\subsection{The dependence of KL on the drift term in a linear model}
\label{app:kltermondrift}

The tightness of the PAC-Bayes generalization guarantee is determined by the size of $\kappa \sum_i \klbin{\lambda_i}{\priorpruneprob_i}$. 
Substituting the optimal $\lambda_i$ from \cref{eq:lambdaoptimal} yields
\begin{equation}
\begin{split}
    \klbin{\lambda_i}{\priorpruneprob_i} =  &-\log (\exp (\eta_i)(1-\priorpruneprob_i) + \priorpruneprob_i) \\
    &+ \eta_i \frac{\exp (\eta)}{\exp (\eta_i) + \frac{\priorpruneprob_i}{1-\priorpruneprob_i}},
    \end{split}
    \label{eq:linearklexact}
\end{equation}

where $\eta_i = \kappa^{-1} \Drift$
We bound \cref{eq:linearklexact} separately for small and large drift $\Drift$.

When $\Drift$, and therefore $\eta_i$, is large positive, then 

$$
\klbin{\lambda_i}{\priorpruneprob_i}  \approx \Drift \priorpruneprob \kappa^{-1}. 
$$
When $\Drift$ is large negative, then 
$$
\klbin{\lambda_i}{\priorpruneprob_i}  \approx \Drift \kappa^{-1} \frac{\priorpruneprob_i}{1- \priorpruneprob_i}  - \log \priorpruneprob_i. 
$$
In both cases, large magnitude of the drift terms results in a linear dependence of the bound on $\Drift$.

When the magnitude of the drift term is small, up to the error introduced by a first order Taylor approximation,  $\lambda_{i,\infty} = \priorpruneprob_i (1 + \kappa^{-1} \Drift (1-\priorpruneprob_i))$. This yields
\begin{equation}
\klbin{\lambda_i}{\priorpruneprob_i}  = \priorpruneprob_i a_i \log a_i - (1- \priorpruneprob_i) b_i \log b_i,
\label{eq:smalldriftklexpression}
\end{equation}
where $a_i = 1-\eta_i \priorpruneprob_i + \eta_i$ and $b_i = 1- \eta_i \priorpruneprob_i$.
Assuming $\eta_i < 1$, we can bound $\log a_i \leq \eta_i (1-\priorpruneprob_i)$ and, similarly, $\log b_i \leq \eta_i \priorpruneprob_i$. Substituting these log bounds to \cref{eq:smalldriftklexpression}, we obtain
\begin{equation}
\klbin{\lambda_i}{\priorpruneprob_i}  \leq \frac{\Drift^2}{\kappa^2} \priorpruneprob_i (1-\priorpruneprob_i).
\end{equation}

In summary, the tightness of the data-dependent PAC-Bayes bound on the generalization error scales with the drift term:
\begin{itemize}
    \item with $|\Drift| \ll 1$, the bound is quadratic in $\Drift$;
    \item with $|\Drift| \gg 1$, the bound is linear in $\Drift$.
\end{itemize}
\subsection{Overparameterized neural networks as linear models}\label{app:NTK}
Recent work by \citet{jacot} has shown that training a depth $L$ neural networks with gradient flow is equivalent to a functional gradient flow with respect to the Neural Tangent Kernel (NTK) defined by
 $  \Theta^L_{t}(x, x') = \nabla_{\theta} f(x, \theta_t) \nabla_{\theta} f(x', \theta_t)^T,$
where $f$ is the network output, $\theta$ are the parameters, and $t$ is the training time. 
In the infinite width limit, the kernel $\Theta^L$ becomes independent of $t$ and training the network with quadratic loss is equivalent to the following linear model
\begin{equation}\label{equation:generalization_formula1}
f_t(x) = f_0(x) + \gamma(x, X_P) (I - e^{-\frac{1}{N} \hat{\Theta}^L t}) (Y_P - f_0(X_P)),
\end{equation}
where $\hat{\Theta}^L = \Theta^L(X_P,X_P)$, $f_0$ is the network output at initialization, and $\gamma(x, X_P) = \Theta^L(x, X_P) (\hat{\Theta}^L )^{-1}$.
Consider a zero init, i.e., $f_0 = 0$, then, in the limit of infinite width, training the network with gradient flow on $S_P$ is equivalent to training the linear model with the feature map given by $\Psi(x) = \{ \Theta^L(x,x')\}_{x'\in S_P}$. Thus, by considering this alternative linear model, all the results on the linear model in this section apply. Note that the most practically relevant setting for pruning is the wide-but-finite DNN setting, where it has been shown that the linear DNN approximation holds around parameter initialization under a certain parameterization or learning rate regime \citep{lee2019wide}.
\section{Experimental Details}
\label{app:experimentaldetails}
\subsection{Networks and datasets} 
Our image classification results consider three tasks: MNIST\citep{lecun-mnisthandwrittendigit-2010}, Fashion-MNIST \citep{xiao2017fashion} and CIFAR-10 \citep{Krizhevsky09learningmultiple}. 
For MNIST we use an MLP with 3 hidden layers of width 1000, like in \citep{baykal2018data}. For Fashion-MNIST we consider the LeNet-5 architecture \citep{Lecun1998lenet}, and for CIFAR-10 we take the ResNet20 architecture \citep{kaimingResnet20}. 
For all datasets we standardize all input images based on the training split, and additionally use randomized crops and horizontal flips for CIFAR
10.
\subsection{Hyperparameters for PAC-Bayes pruning (PBP)}\label{sec:hyperparameters}
Throughout all stages of training, we optimize using SGD with momentum 0.9, batch size 128.
Below we provide the details for the individual stages from  \cref{sec:main_exps}.

\paragraph{Stage 1. Pre-training the prior.}
 We used fixed learning rates of 0.01 in MNIST \& Fashion-MNIST and 0.1 in CIFAR-10. We trained for 30 epochs on MNIST, 70 epochs on Fashion-MNIST and 120 epochs on CIFAR-10 respectively. For CIFAR-10 we also decayed the learning rate by 0.1 at 50\% and 75\% of training.

\paragraph{Stage 2. Fine-tuning the stochastic prior.}
In the experiments presented in \cref{sec:main_exps}, we pruned the deterministic prior using MBP, according to the desired sparsity level. In order to maintain the same level of sparsity in our stochastic prior, we set $\priorpruneprob$ using the block isotropic prior introduced in \cref{app:initializing_pruning_probs}, setting $\epsilon=0.0001$ in \cref{eq:blockisotropicprior} in the end after initially tuning this hyperparameter. We train for 30 epochs on MNIST, 30 epochs on Fashion-MNIST and 60 epochs on CIFAR-10. We used the same learning rates as in Stage 1 for all tasks. Again on CIFAR-10, we decay the learning rate by 0.1 twice during the course of training in both Stage 2 and Stage 3. On all tasks we set the per weight variance $\sigma_i^2$ to be constant over parameters, and tune this constant on a log-uniform (base $e$) scale between $[-15, -3]$. Whenever we tune hyperparameters, we always seek to minimize the PAC-Bayes generalization bound on error. For both $\sigma_i$ and posterior learning rate (next paragraph), we tuned only on a finite grid of values, and by a union bound argument this adds only a negligible penalty on top of our PAC-Bayes bound.
\paragraph{Stage 3.  Training the posterior.}
We train for the same number of epochs as in Stage 2 in all tasks, using an initial learning rate that is tuned on a log-uniform scale between $[-4, -1], [-7, -4], [-5, 2]$ respectively for MNIST, Fashion-MNIST and CIFAR-10. In our experiments, we found the lowest generalization bounds to arise when pruning probabilities were set to extreme values, (close to 0 or 1), due to the performance-risk trade off as described in \cref{sec:tradeoffs}.  We set $\delta=0.05$ when calculating our generalization bounds.
In \cref{sec:main_exps}, our goal was to demonstrate that we can obtain good PAC-Bayes generalization bounds on small/moderate scale image classification tasks using our stochastic pruning framework. It turns out we can obtain surprisingly good results without training posterior pruning probabilities (see results in  \cref{fig:EvS} ).

\subsection{Hyperparameters choice for Probabilistic Fine-Tuning(PFT)}
PFT uses the same hyperparameters as stated in \cref{sec:hyperparameters}. 
More precisely, for training the weights of the unpruned deterministic classifier, we use the hyperparameters from Stage 1.
Then we find a randomized sparse network by performing PFT using the hyperparameters from Stage 2.
Finally, we fine-tune the weights of the corresponding deterministic sparse classifier using the hyperparameters from Stage 1.

\section{Additional Empirical Results}
\label{app:additionalresults}
Here we present some additional experiments complementing our results in the main text on PFT and PBP. 
\subsection{Probabilistic Fine-Tuning (PFT)}\label{app:pft}
\begin{figure}
    \centering
    \includegraphics[width=.95\linewidth]{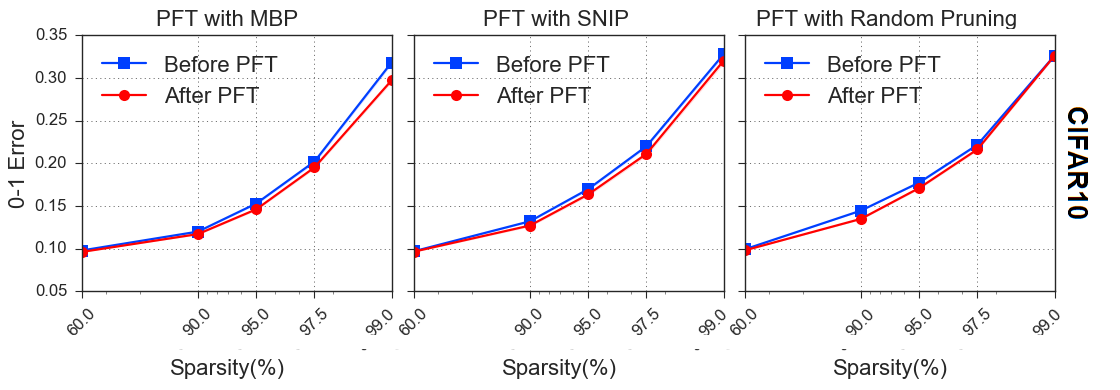}
    \caption{\small{Impact of the application of PFT on standard One Shot Pruning methods (MBP, SNIP) and Random Pruning. CIFAR-10 with ResNet20.}}
    \label{fig:pft_results_CIFAR10}
\end{figure}

This section complements the results on PFT shown in \cref{fig:pft_results}. \cref{fig:pft_results_CIFAR10} shows full results for CIFAR-10 experiments with ResNet20.

\subsection{PFT at initialization}

\begin{figure*}[t]
    \centering
    \includegraphics[width=.8\linewidth]{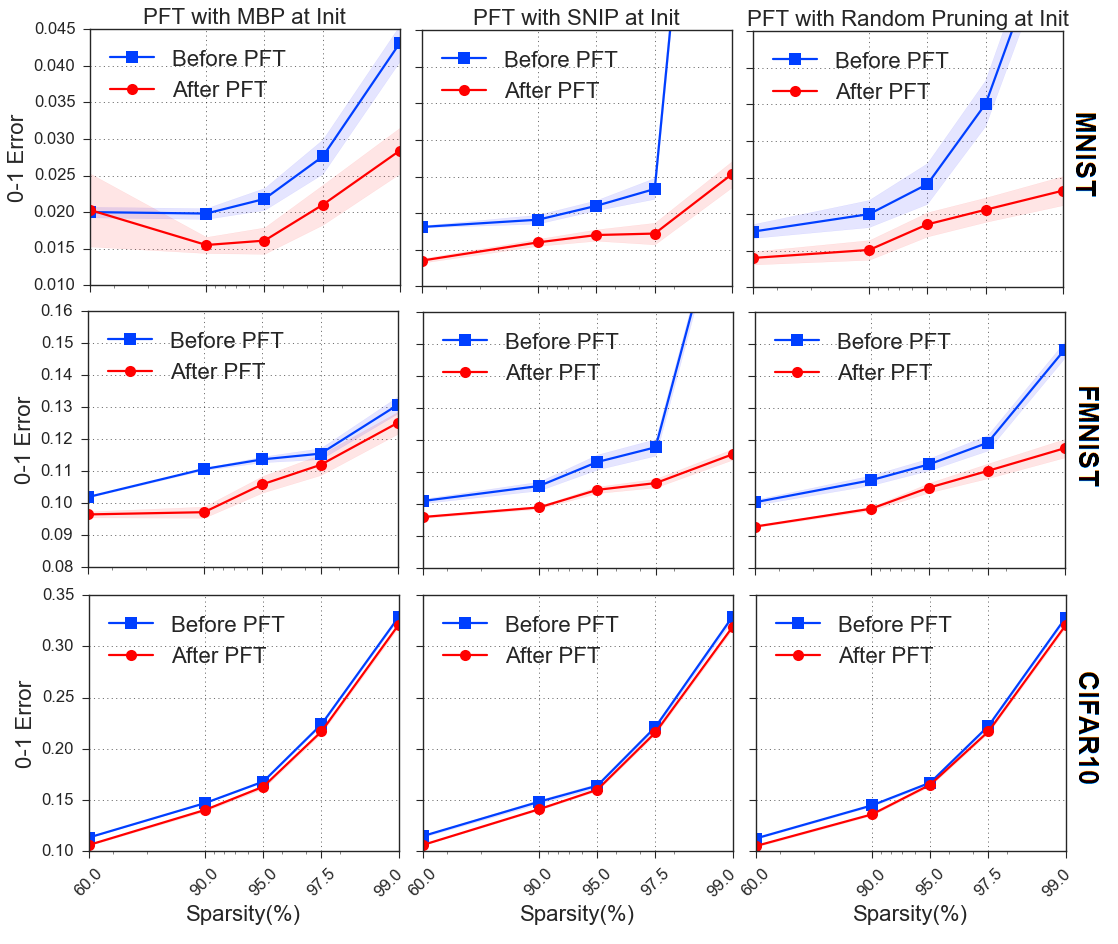}
    \caption{\small{Impact of the application of PFT on standard One Shot Pruning methods (MBP, SNIP) and Random Pruning at Initialization. MNIST and FMNIST with (MLP 1000x3) and CIFAR-10 with ResNet20.}}
    \label{fig:pft_results_pft_init}
\end{figure*}

We run multiple experiments with PFT at initialization, where we skip the pre-training of an unpruned network phase (Step 1).
Instead, we optimize a random pruning mask from the weigths at initialization, in a similar fashion to \citep{lee2018snip, wangGRASP, hayouPruning, frankle2020missing}. 
The goal is to assess whether PFT improves upon existing methods of One-Shot Pruning at initialization, namely Magnitude Pruning, SNIP \citep{lee2018snip} and Random Pruning. 
\cref{fig:pft_results_pft_init} shows that PFT improves the performance of existing OSP methods at initialization in almost all the settings we test.
We hypothesize that this is due to the fine-tuning step finding masks that better adapt to the data.

We emphasize, however, that this additional benefit of fine-tuning is not that surprising:
OSP methods use one forward/backward pass through the network and the training data to create the mask; our method requires multiple forward/backward passes to fine-tune the probabilities.
However, it is interesting that even local optimization starting from the weights at initialization can boost the performance of pruning-at-initialization methods. 

\subsection{Strong Lottery Ticket Hypothesis and PFT}

\begin{figure*}[t]
    \centering
    \includegraphics[width=.4
    \linewidth]{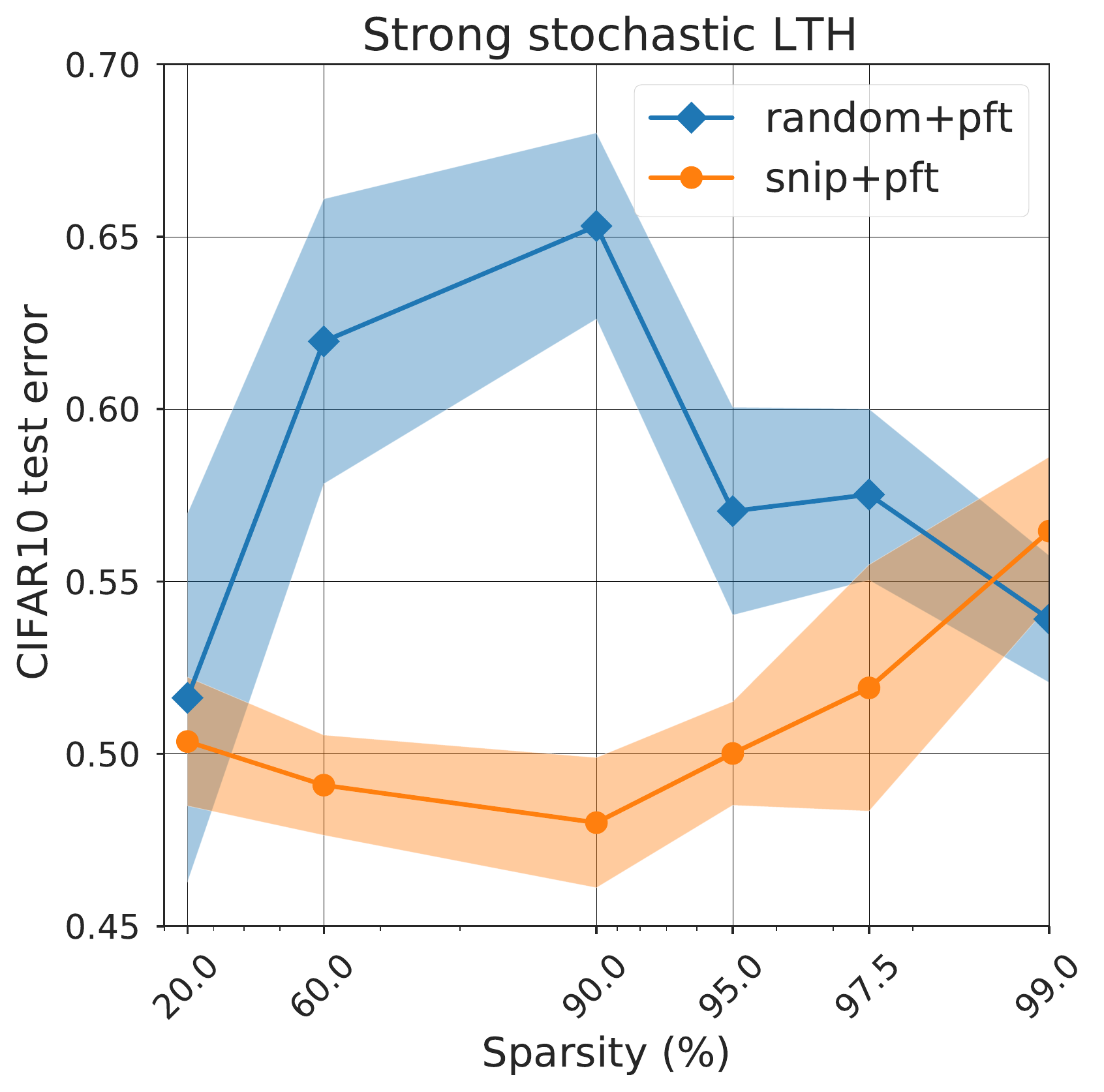}
    \caption{\small{Strong stochastic Lottery Ticket Hypothesis test errors on CIFAR-10 with ResNet20 for PFT with random or SNIP initialisation of stochastic mask. Plotted intervals correspond to 95\% confidence from 10 independent samples from the trained stochastic mask.}}\label{fig:stoch_strong_lth}
\end{figure*}

Strong Lottery ticket hypothesis (Strong LTH) states that at initialization there exists a sparse subnetwork that achieves competitive accuracy, if the network is sufficiently overparameterized.
We apply PFT at initialization without any weight training before or after pruning.
\cref{fig:stoch_strong_lth} presents the results on 
CIFAR-10, showing that PFT finds random pruning masks that, when applied to untrained weights, yield a network obtaining nontrivial test error.
Here for a given sparsity we initialized the stochastic mask with either isotropic stochastic mask (PFT starting from a random mask) or a block isotropic stochastic mask (PFT starting from SNIP mask) set with $\epsilon=0.01$ in \cref{eq:blockisotropicprior}, and trained the masks for 100 epochs with learning rate 0.5. 
Note in both cases without PFT, the stochastic mask would obtain 90\% error on untrained weights.

\newpage
\section{PAC-Bayes Bound Optimization}

\subsection{Performance--risk bound tightness trade-off}
\label{sec:tradeoffs} 
\begin{figure}[tbh]
    \centering
    \includegraphics[width=0.8\linewidth]{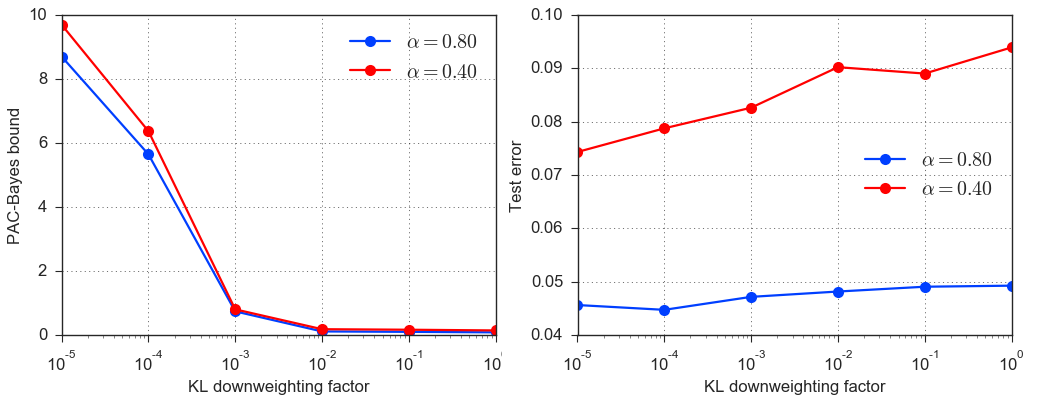}
    \caption{Performance--Bound trade-off with a 1000x3 MLP trained on MNIST, starting from Random Pruning.}
    \label{fig:performance_bound_tradeoff}
\end{figure}

The KL divergence term plays a crucial role in the PAC-Bayes bound optimization. 
To capture the impact of the KL term, we run multiple experiments where we down-weigh the KL term in the bound, i.e., we multiply the KL term by a factor $d$ and optimize the new bound. Intuitively, since the KL term acts as a penalty forcing the posterior $Q$ to remain close to the prior $P$, we expect the test error of the posterior random classifier $Q$ to improve as we down-weigh the KL term. On the other hand, when we optimize the bound with the down-weighed KL term, we expect the real bound (i.e., without down-weighing) to become worse as $Q$ is able to move away from $P$. 
This trade-off is illustrated in \cref{fig:performance_bound_tradeoff}.

\subsection{Overall Entropy versus Error Bound}

\begin{figure*}[htp]
     \centering
     \begin{subfigure}[b]{0.4\textwidth}
         \centering
         \includegraphics[width=\textwidth]{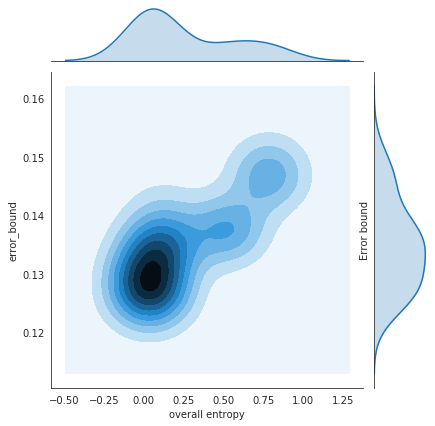}
         \label{fig:kde_plot_entropy_errorbound}
     \end{subfigure}
     \begin{subfigure}[b]{0.4\textwidth}
         \centering
         \includegraphics[width=\textwidth]{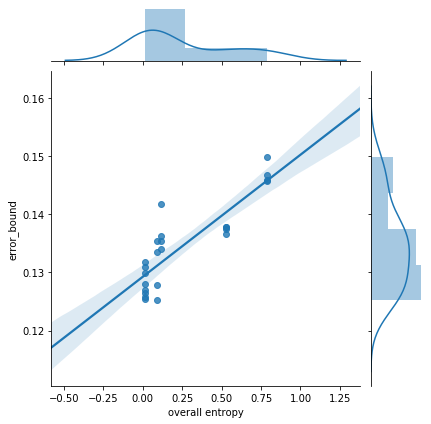}
         \label{fig:reg_plot_entropy_errorbound}
     \end{subfigure}
     \hfill
        \caption{Overall stochastic pruning mask entropy vs PAC-Bayes bound on error with (MLP 1000x3) on FMNIST}
        \label{fig:entropy_vs_error_bound}
\end{figure*}

The binary entropy of the Bernoulli distribution $\mathcal{B}(p)$ is defined by $\ent(p) = -p \log_2(p) - (1-p) \log_2(1-p) $. The binary entropy measures the randomness of Bernoulli variables, it ranges from 0 for $p=0$ or $p=1$, to $1$ for $p=1/2$ (intuitively, when $p=50\%$, the Bernoulli variable is completely random and does not favor one side to the other). In our experiments, we measured the overall entropy of our optimized stochastic pruning masks by taking an average over per-weight-entropy.
\cref{fig:entropy_vs_error_bound} shows how the optimized PAC-Bayes bound  on error correlates with entropy. 
We see a clear trade-off between the tightness of the bound and the overall entropy; as the overall entropy increases, the probabilistic mask changes more significantly across network samples.

\subsection{Evolution of Prior and Posterior Probabilities during training}
\begin{figure*}[htp]
     \centering
     \begin{subfigure}[b]{0.3\textwidth}
         \centering
         \includegraphics[width=\textwidth]{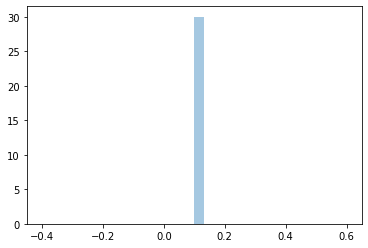}
         \caption{Init Prior}
         \label{fig:output_layer_probs_init_hist}
     \end{subfigure}
     \begin{subfigure}[b]{0.3\textwidth}
         \centering
         \includegraphics[width=\textwidth]{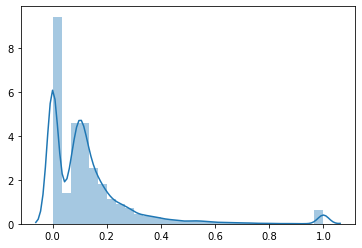}
         \caption{Init Posterior}
         \label{fig:output_layer_probs_posterior_init_hist}
     \end{subfigure}
     \begin{subfigure}[b]{0.3\textwidth}
         \centering
         \includegraphics[width=\textwidth]{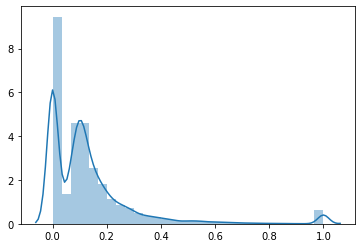}
         \caption{Final Posterior}
         \label{fig:output_layer_probs_posterior_final_hist}
     \end{subfigure}
     \hfill
        \caption{Probabilities Histogram of the output layer in a (MLP 1000x3) on MNIST, starting from Random Pruning}
        \label{fig:hist_probs}
\end{figure*}

The prior and posterior pruning probabilities are optimized in the fine-tuning stage and posterior training (Stages 2, 3 in \cref{sec:stochastic_pruning}), respectively. 
The pruning probabilities are optimized to minimize the empirical risk in Stage 2 and PAC-Bayes bound in Stage 3. 
Starting from a randomly pruned classifier with sparsity $s=90\%$, i.e., setting $\priorpruneprob_i = 10\%$ for all $i$'s, through a histogram we monitored how the probabilities change after Stage 2 and Stage 3. 
\cref{fig:hist_probs} shows the histogram of pruning probabilities of the output layer of a 3 layer MLP with 1000 neurons per layer, trained on MNIST. 
At initialization, the pruning probabilities are concentrated at $10\%$ (random pruning, \cref{fig:output_layer_probs_init_hist}). After the fine-tuning phase (stage 2), \cref{fig:output_layer_probs_posterior_init_hist} shows that the probabilities spread out over the interval $(0,1)$, with some converging to $1$ and others to $0$, with the bulk remaining in $(0,1)$. This suggests that the empirical risk of the prior is not minimized by a deterministic mask. In the posterior training phase (Stage 3), from \cref{fig:output_layer_probs_posterior_final_hist} (c) we see that the probabilities do not deviate from Stage 2.
This is due to the KL term acting as a regularizer and forcing the posterior to stay close to the prior.

\begin{figure*}[htp]
     \centering
     \begin{subfigure}[b]{0.6\textwidth}
         \centering
         \includegraphics[width=\textwidth]{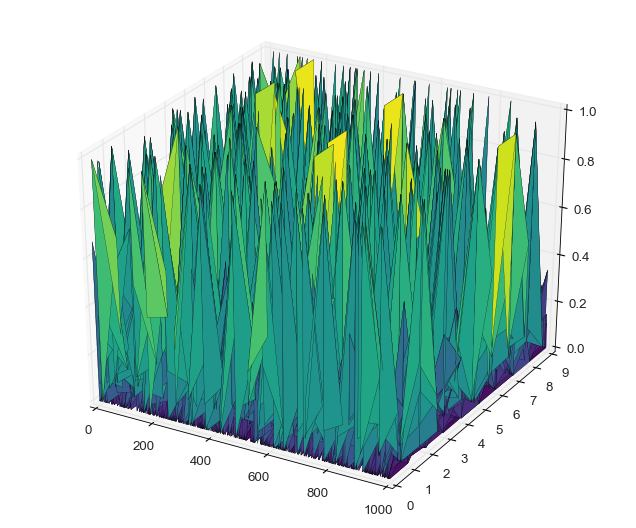}
     \end{subfigure}
     \hfill
        \caption{3D plot of Posterior Probabilities in the output layer in a (MLP 1000x3) on MNIST, starting from Random Pruning.}
        \label{fig:hist_probs_3D}
\end{figure*}

\begin{figure*}[htp]
     \centering
     \begin{subfigure}[b]{0.7\textwidth}
         \centering
         \includegraphics[width=\textwidth]{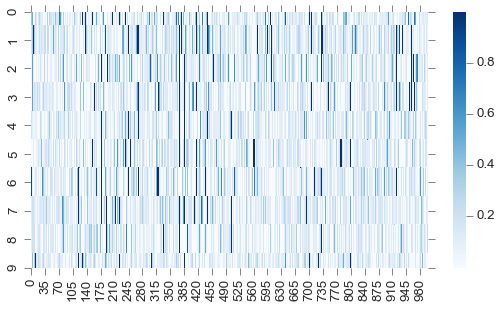}
     \end{subfigure}
     \hfill
        \caption{Heatmap plot of Posterior Probabilities in the output layer in a (MLP 1000x3) on MNIST, starting from Random Pruning.}
        \label{fig:hist_probs_heatmap}
\end{figure*}

To visualize how the posterior pruning probabilities are distributed across the weights, 
in \cref{fig:hist_probs_3D}
we plot a 3D map of the posterior probabilities in the output layer (with dimensions 10 by 1000). The $z$-axis plots $\lambda_i$'s, i.e., the probability that the weights is kept. 
In \cref{fig:hist_probs_heatmap}, we further plot a heatmap of $\lambda_i$'s for the output layer. 
Both plots show that high values of $\lambda_i$  are spread out across the output layer.

\begin{figure*}[htp]
     \centering
     \begin{subfigure}[b]{0.7\textwidth}
         \centering
         \includegraphics[width=\textwidth]{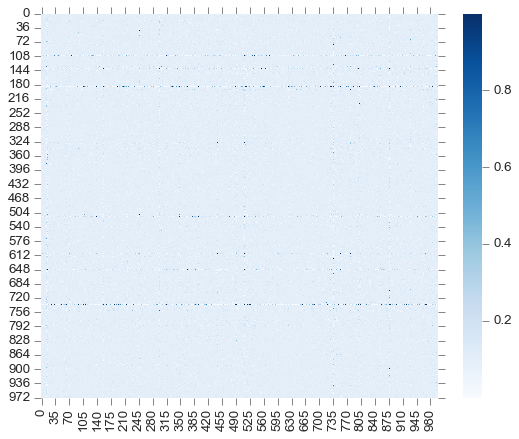}
         \label{fig:mid_layer_probs_posterior_final_heatmap}
     \end{subfigure}
     \hfill
        \caption{Heatmap plot of Posterior Probabilities in the mid layer in a (MLP 1000x3) on MNIST, starting from Random Pruning}
        \label{fig:hist_probs_heatmap_mid_layer}
\end{figure*}

\newpage
\subsection{PairPlot analysis for PBP}
\begin{figure*}[htp]
     \centering
     \begin{subfigure}[b]{.98\textwidth}
         \centering
         \includegraphics[width=\textwidth]{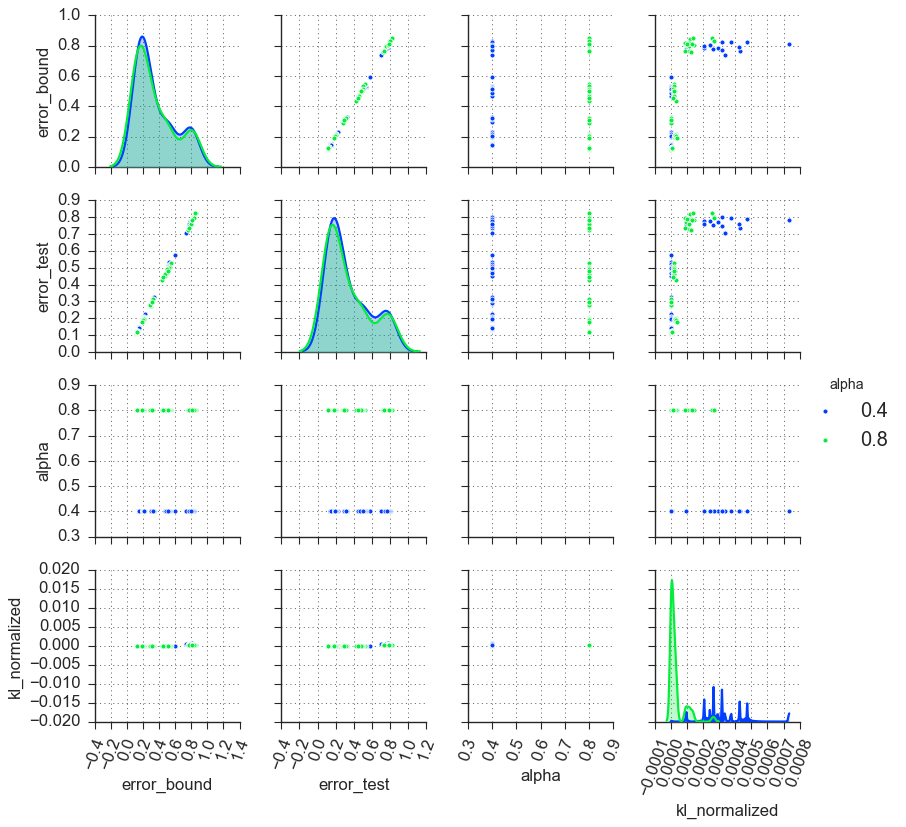}
     \end{subfigure}
     \hfill
        \caption{Pairplot of our experiments on CIFAR-10 with ResNet20,  with Magnitude Based Pruning}
        \label{fig:pairplot_CIFAR10}
\end{figure*}

Here we seek to evaluate how various quantities appearing in PBP optimization interact with each other.
Pairplot analysis is a valuable tool to depict such interactions. \cref{fig:pairplot_CIFAR10} shows a pairplot of 4 quantities: PAC-Bayes bound , test error of the posterior $Q$, the fraction of the weight used to train the prior $P$ , and the normalized KL divergence. 
The first noticeable effect is how the choice of alpha impacts the normalized KL divergence. Indeed, the histogram of the normalized KL divergence shows two bulks of values, each corresponding to a value of alpha. Overall, larger alpha tends to yield smaller KL divergence. This can be attributed to the fact that using more data to train the prior would allow the posterior not to move far from the prior in order to find a good trade-off between the KL term and the empirical risk.

Another interesting observation is the almost perfect linear relationship between the posterior test error and the PAC-Bayes bound. This suggests that the KL term is usually negligible in comparison with posterior risk, showing that posterior training favors localized optimization (i.e., concentrated around the prior).  
This is also confirmed by the plot of the PAC-Bayes bound as a function of the normalized KL divergence; on average, high KL divergence yields worse PAC-Bayes bounds, confirming that fact that PBP optimizes posterior weights and pruning probabilities in the neighborhood of the prior.
\newpage

\subsection{Is the deterministic mask really deterministic?}
\begin{figure*}[htp]
     \centering
     \begin{subfigure}[b]{0.7\textwidth}
         \centering
         \includegraphics[width=\textwidth]{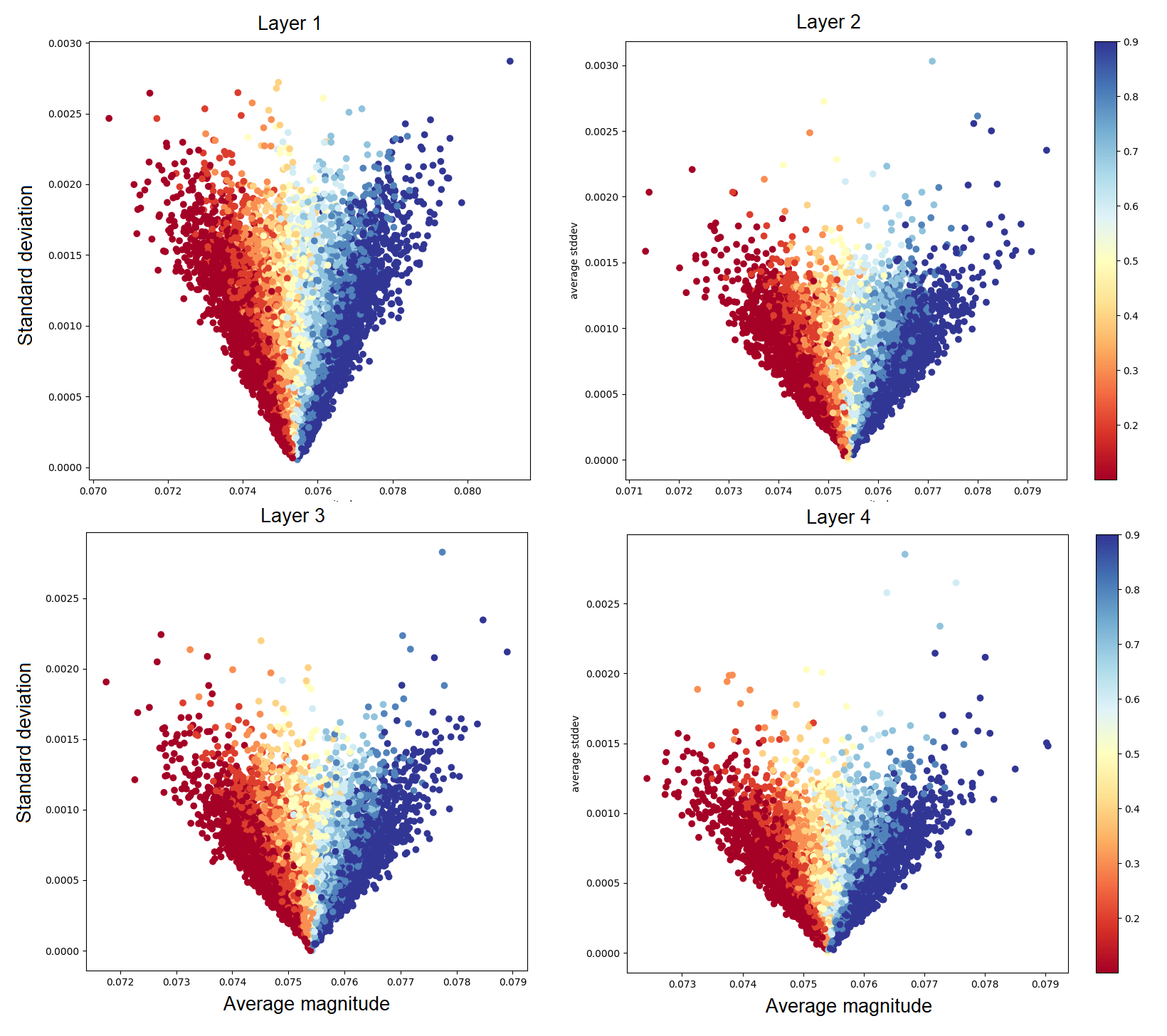}
     \end{subfigure}
        \caption{Plot showing the weights that change the mask (with Magnitude pruning) when data is reshuffled. X--axis is the average magnitude of the weight and Y--axis is its standard deviation. The architecture is MLP 1000x5 trained on MNIST. We only show the middle layers. The color scale represents the probability that the weights is unpruned (in 30 runs).}
        \label{fig:changing_weights_mlp}
\end{figure*}

\begin{figure*}[htp]
     \centering
     \begin{subfigure}[b]{0.4\textwidth}
         \centering
         \includegraphics[width=\textwidth]{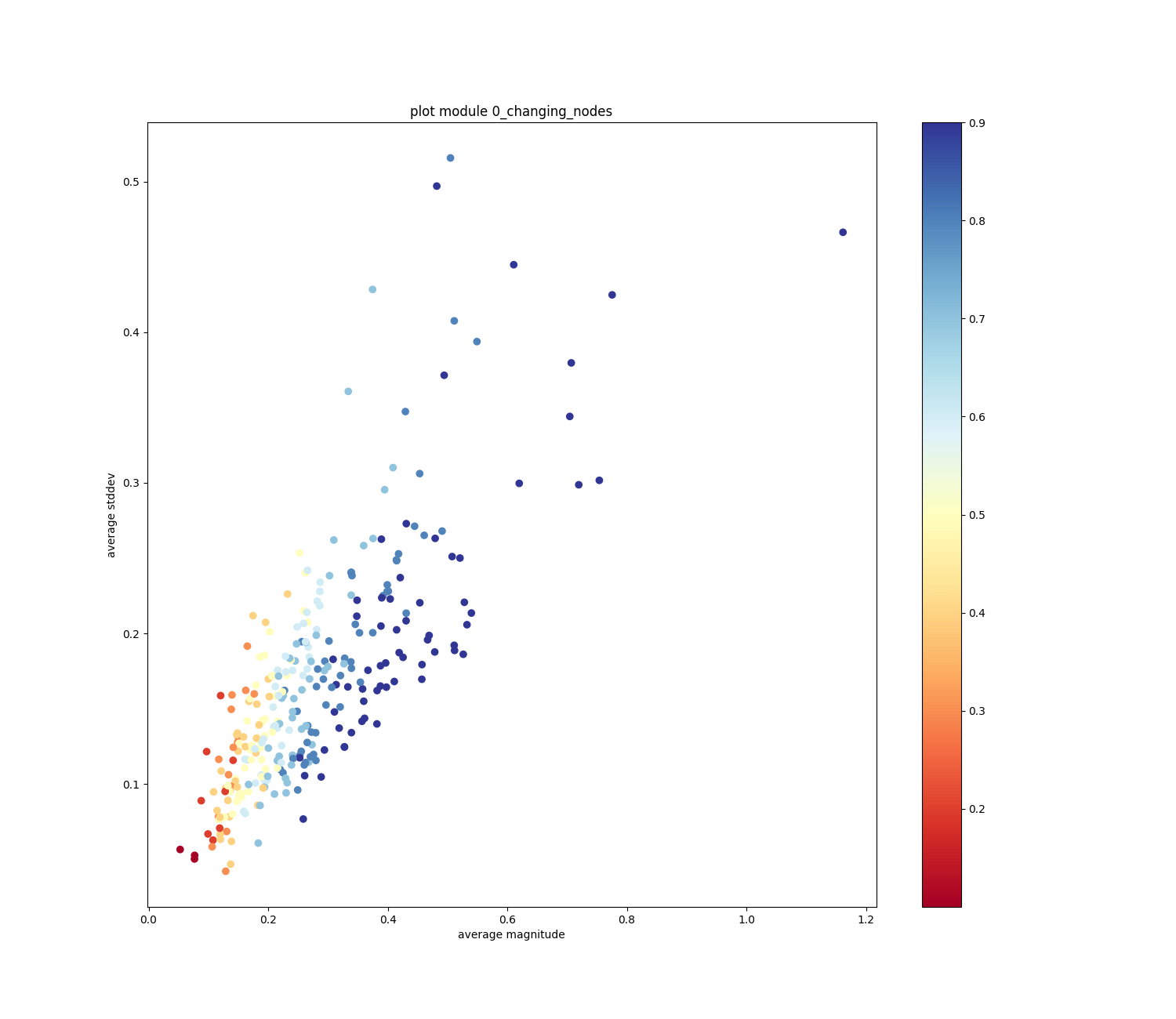}
     \end{subfigure}
     \begin{subfigure}[b]{0.4\textwidth}
         \centering
         \includegraphics[width=\textwidth]{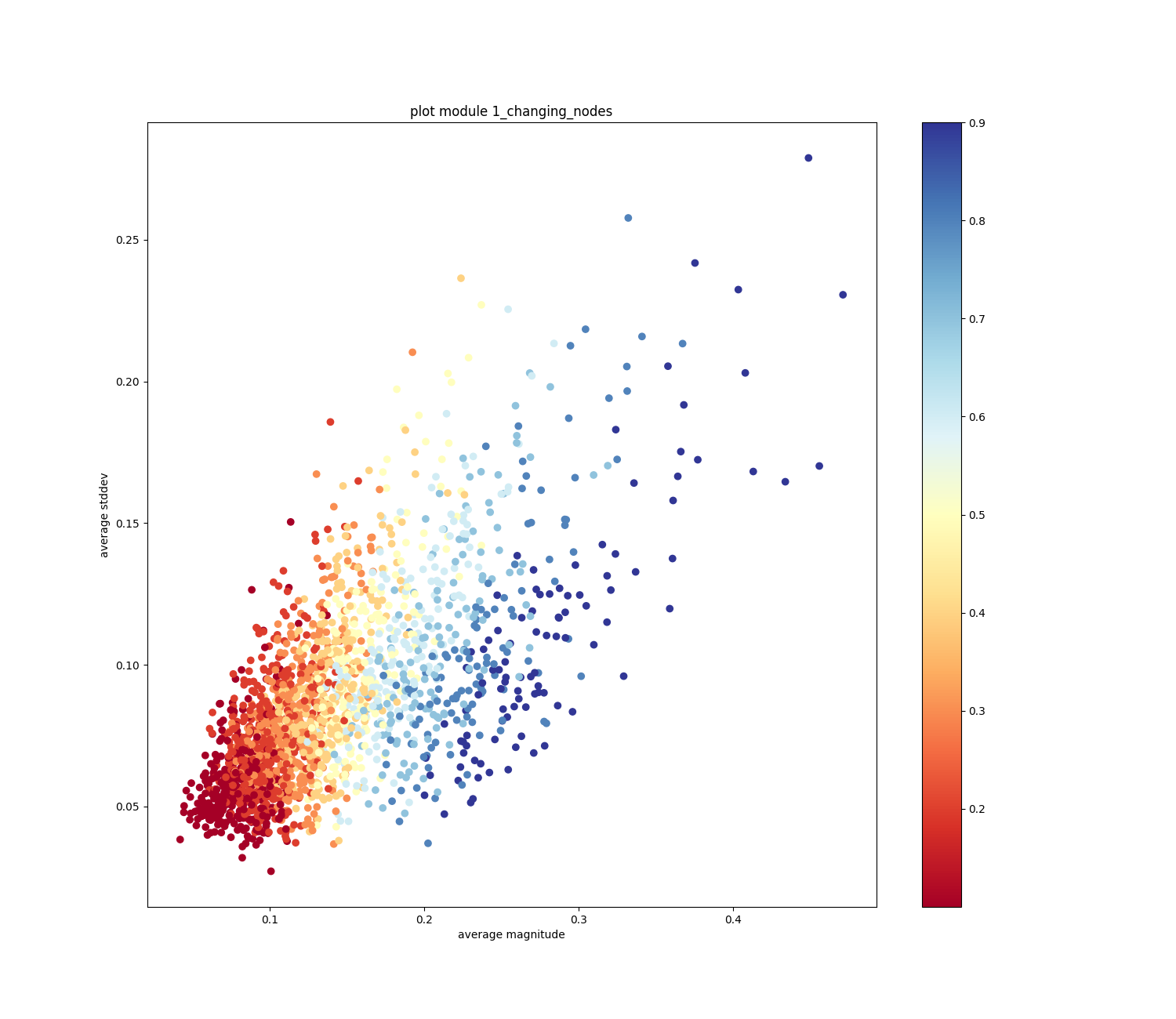}
     \end{subfigure}
     \begin{subfigure}[b]{0.4\textwidth}
         \centering
         \includegraphics[width=\textwidth]{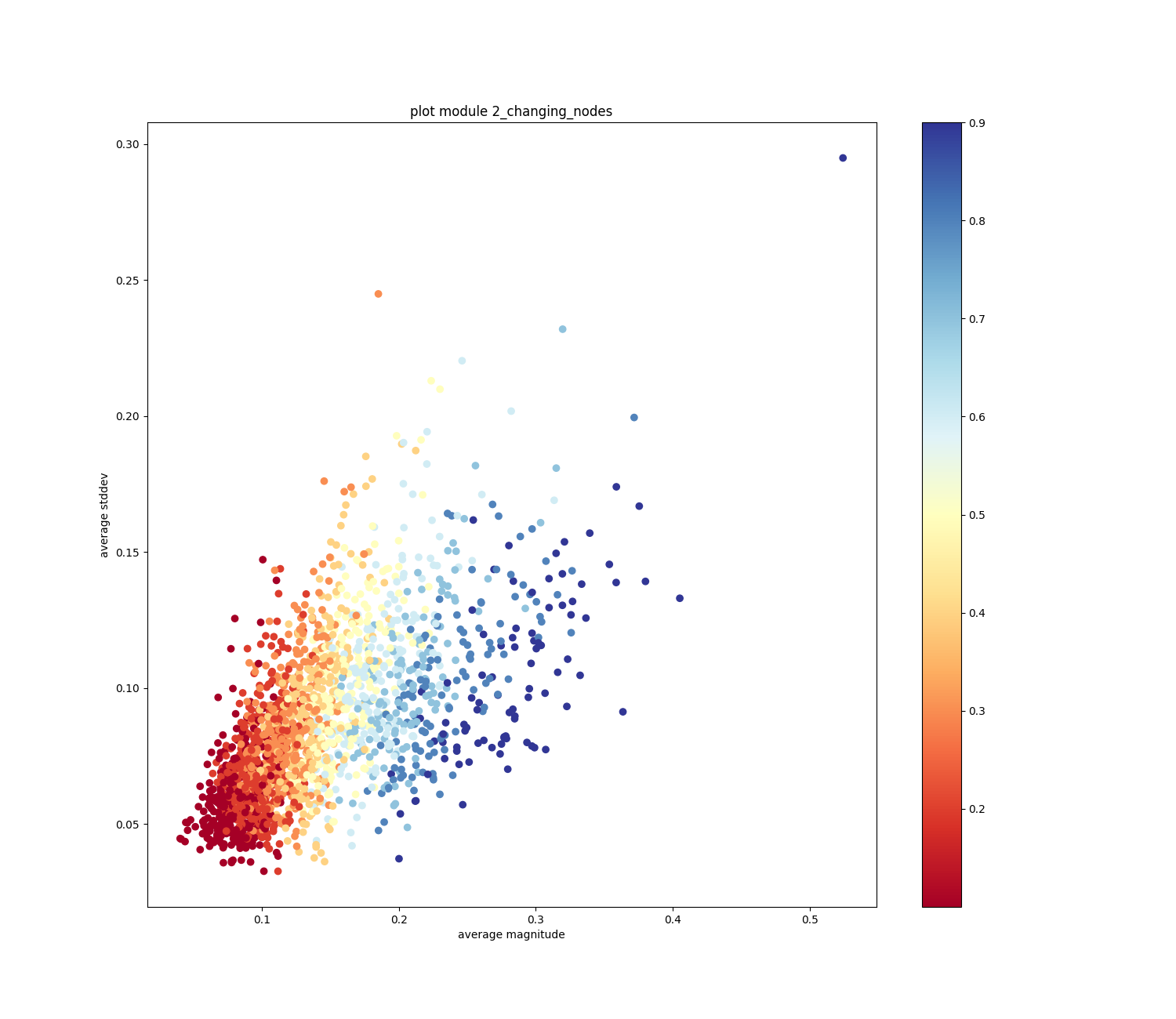}
     \end{subfigure}
     \begin{subfigure}[b]{0.4\textwidth}
         \centering
         \includegraphics[width=\textwidth]{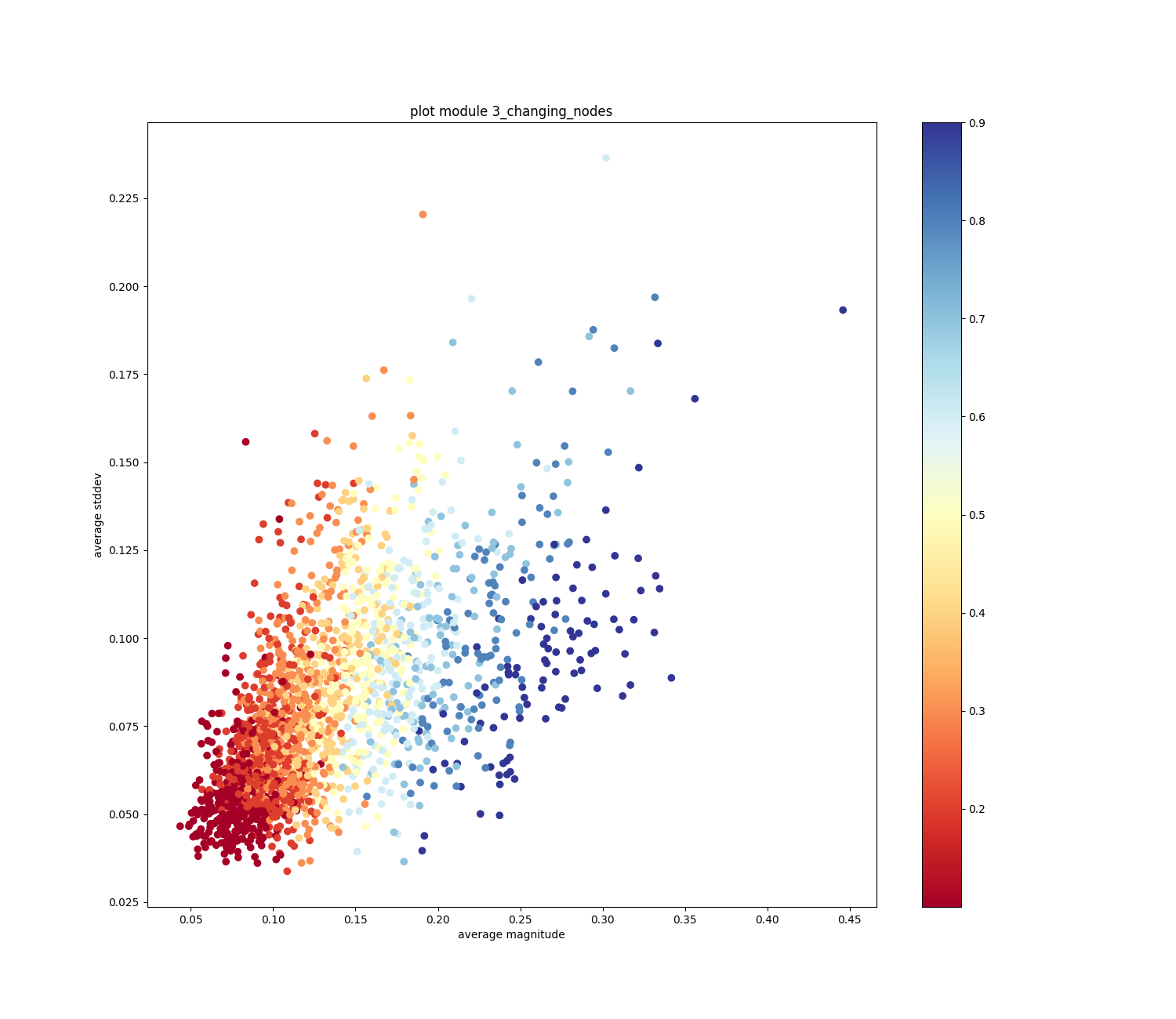}
     \end{subfigure}
     \begin{subfigure}[b]{0.4\textwidth}
         \centering
         \includegraphics[width=\textwidth]{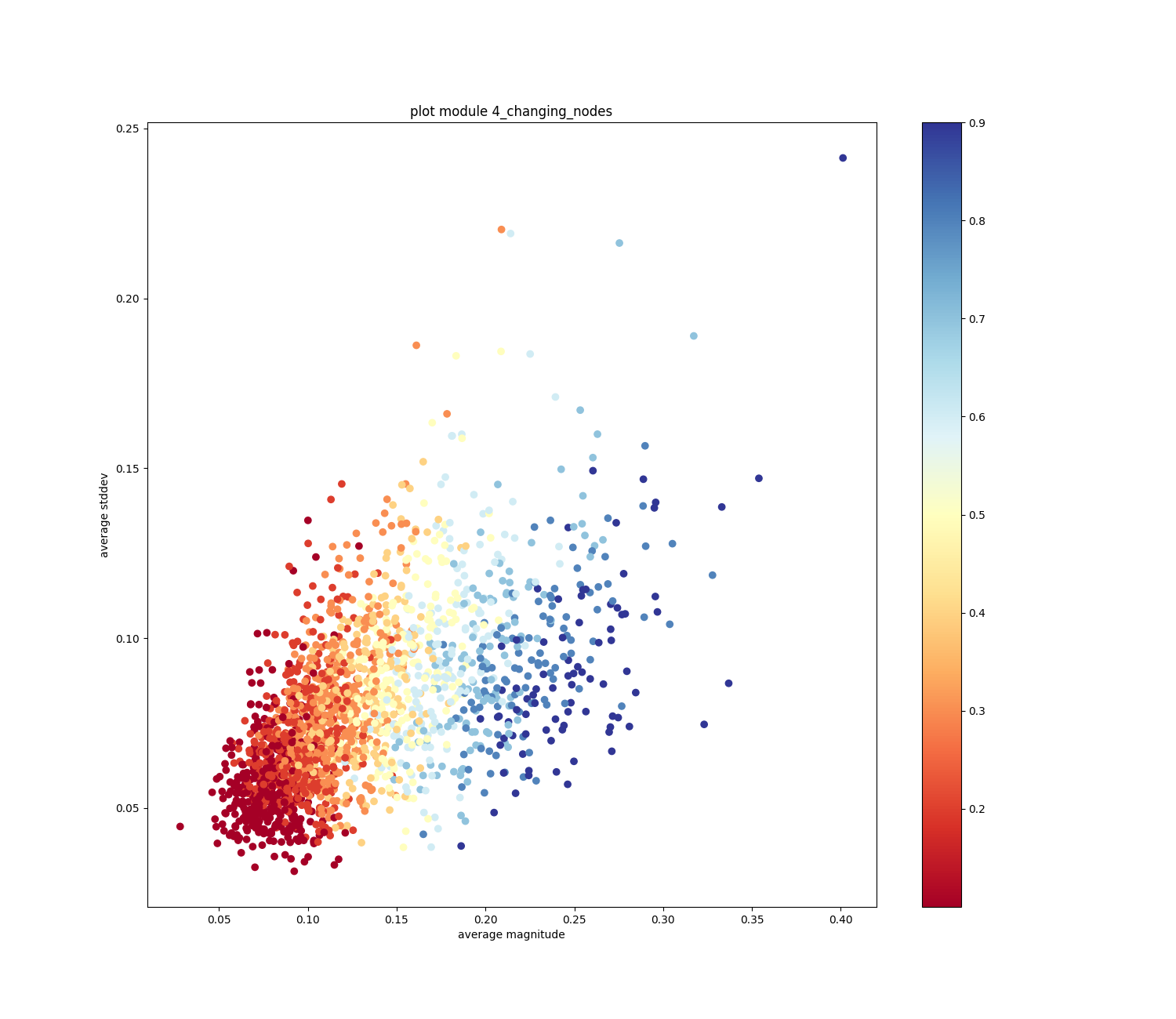}
     \end{subfigure}
     \begin{subfigure}[b]{0.4\textwidth}
         \centering
         \includegraphics[width=\textwidth]{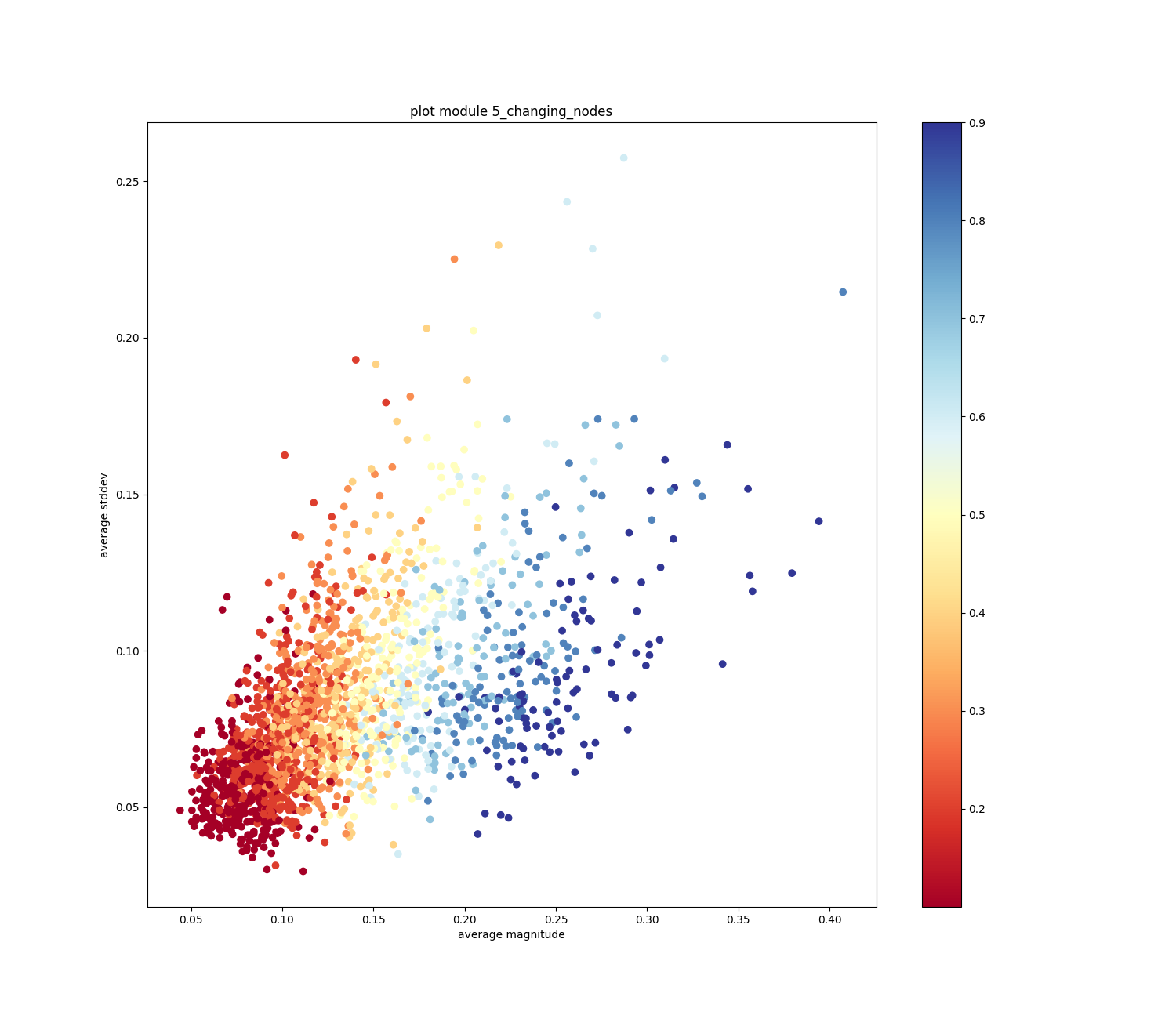}
     \end{subfigure}

    \caption{Plot showing the weights that change the mask (with Magnitude pruning) when data is reshuffled. X--axis is the average magnitude of the weight and Y--axis is its standard deviation. The architecture is ResNet20 trained on CIFAR-10.(first 6 layers)}
    \label{fig:changing_weights_CIFAR10}
\end{figure*}

It is well known that with standard pruning methods, the resulting sparse architecture depends on the initialization of the weights and data reshuffling seed. The dependence on the initialization is natural and expected: a weight that was initialized to be very small compared to others will likely remain small after training, especially in sufficiently overparameterized neural networks trained with small learning rates. 
However, the dependence of the mask on the data shuffling seed is interesting and suggests that there may be no single optimal sparse architecture, and a randomized mask might be better. 
We train a network multiple times with a fixed initialization. Each run has a different data shuffling seed.
We keep track of the probability of the weight being pruned.
\cref{fig:changing_weights_mlp} illustrates our result on an MLP with 5 layers. Here we only plot the weights that have an empirical pruning probability in $(0,1)$.
We show the standard deviation of these weights as a function of their average magnitude across the runs. 
We provide similar plots for CIFAR-10 with ResNet20 in \cref{fig:changing_weights_CIFAR10}.

\begin{figure*}[htp]
     \centering
     \begin{subfigure}[b]{0.4\textwidth}
         \centering
         \includegraphics[width=\textwidth]{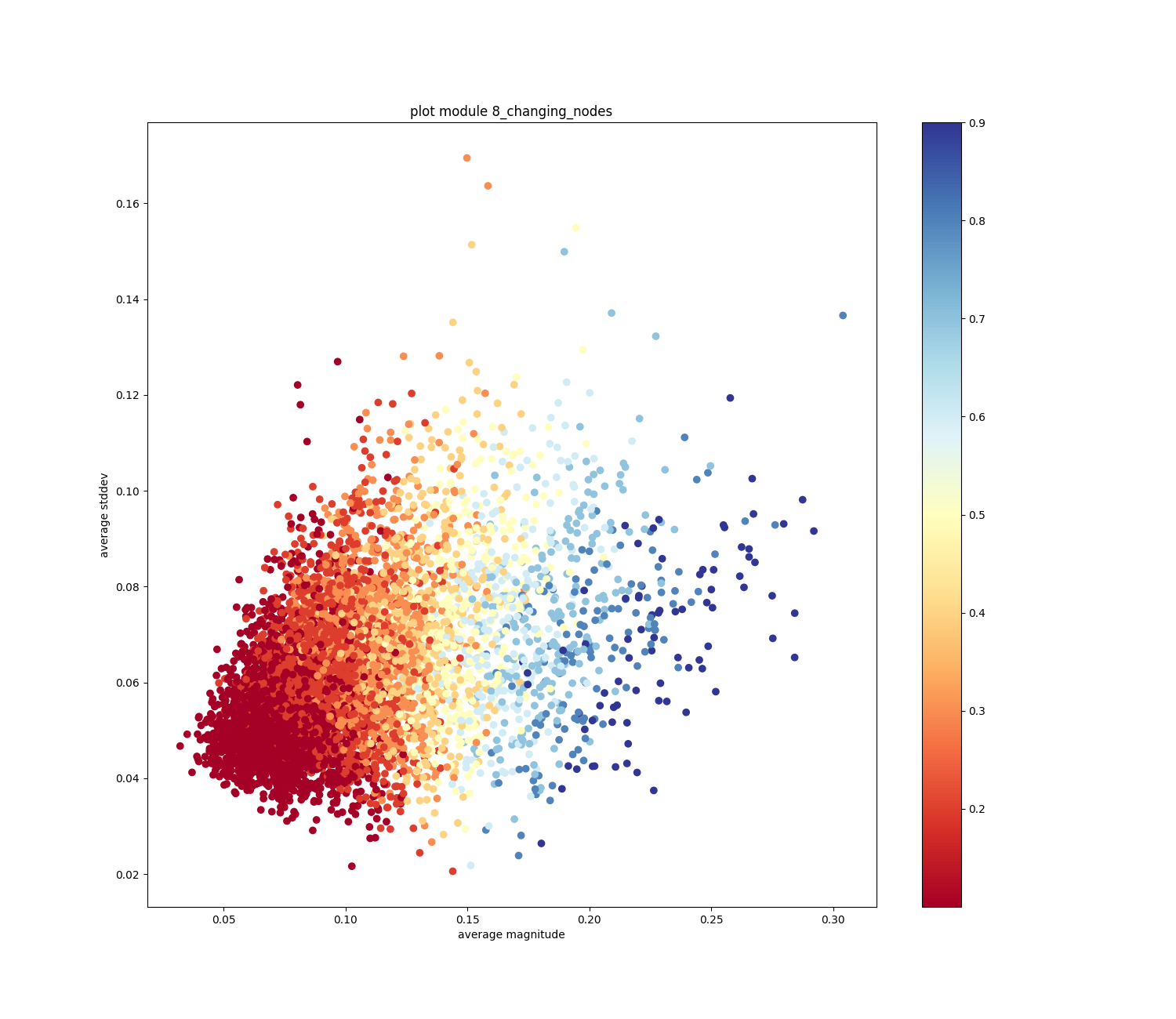}
     \end{subfigure}
     \begin{subfigure}[b]{0.4\textwidth}
         \centering
         \includegraphics[width=\textwidth]{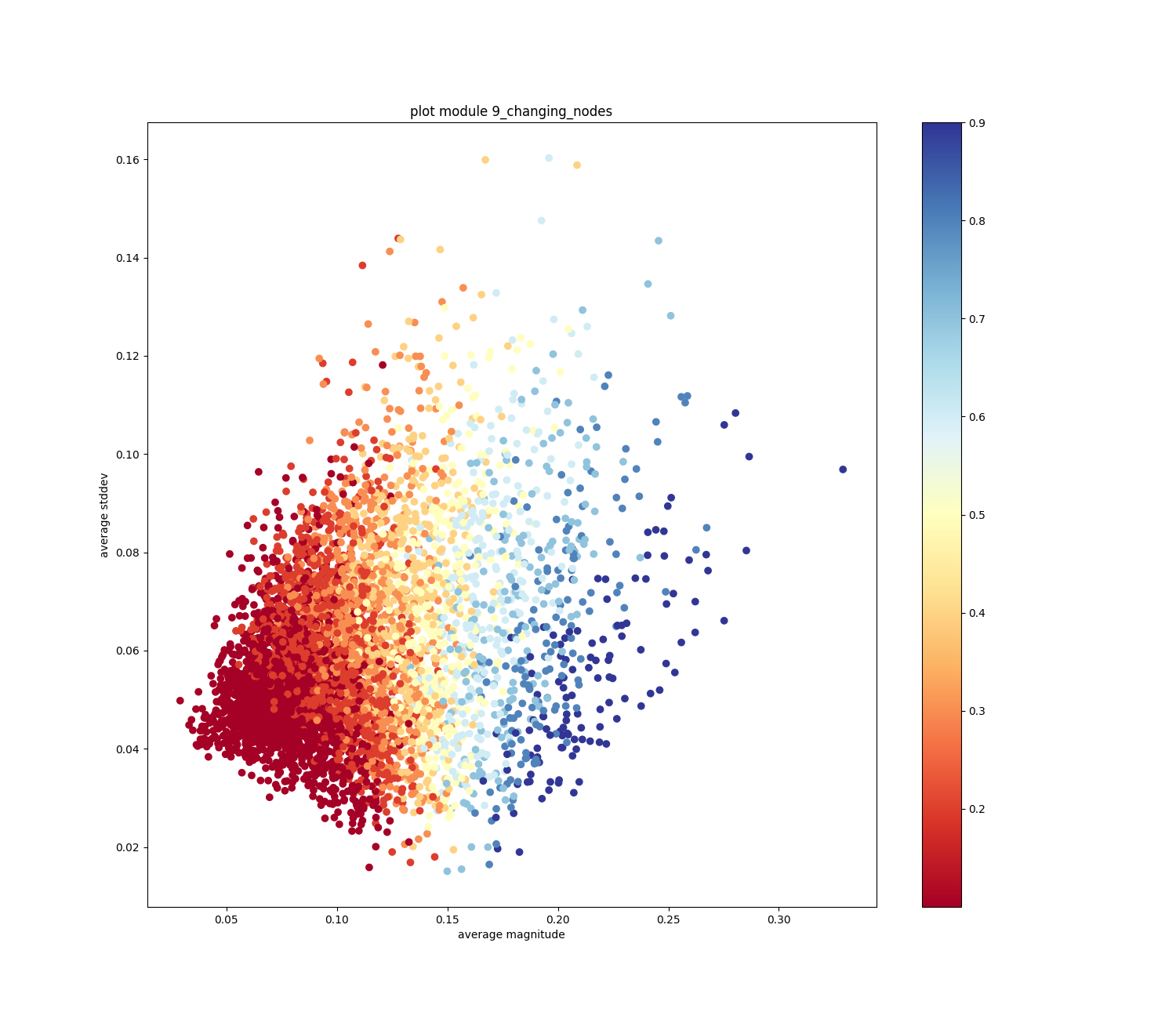}
     \end{subfigure}
     \begin{subfigure}[b]{0.4\textwidth}
         \centering
         \includegraphics[width=\textwidth]{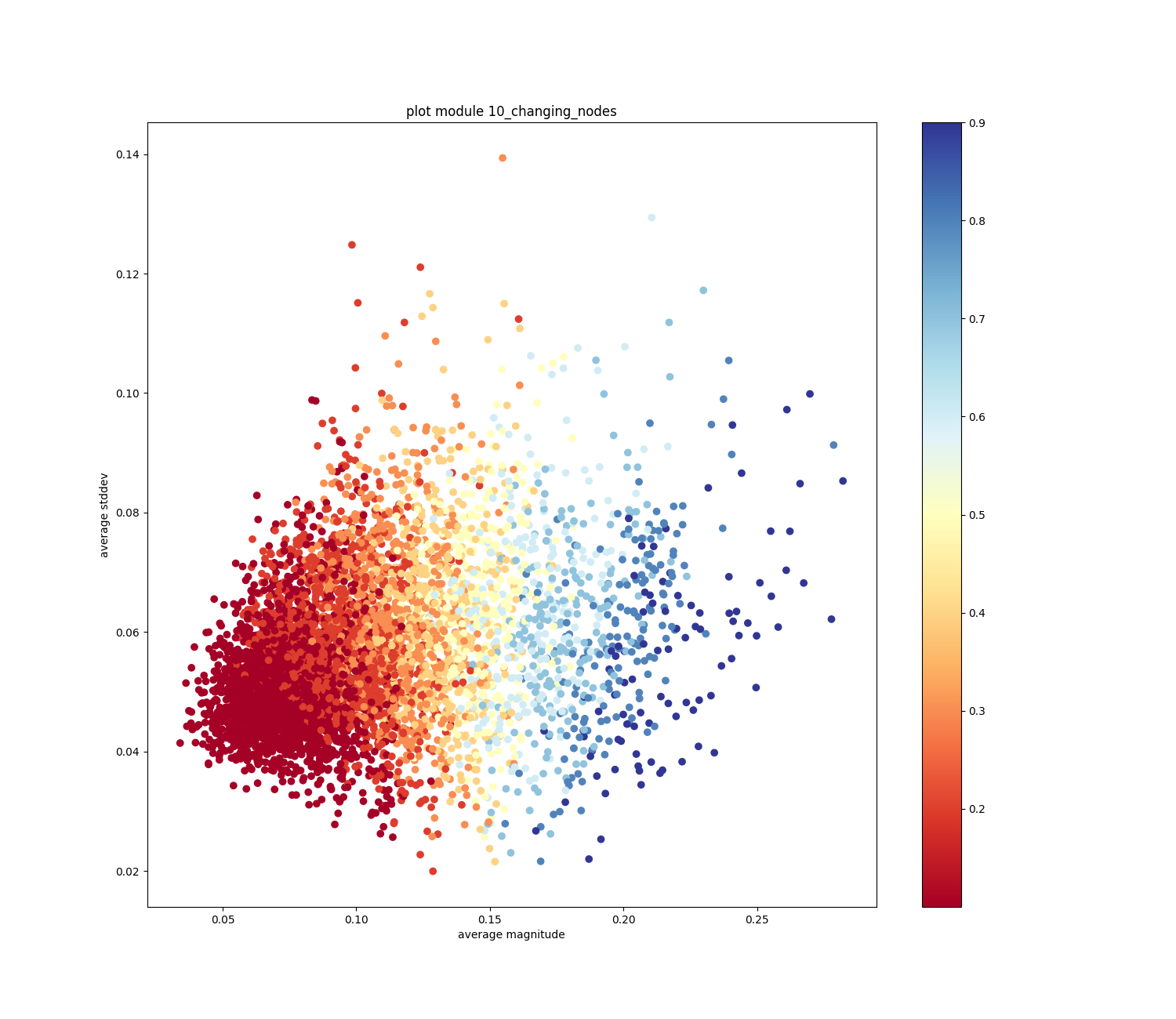}
     \end{subfigure}
     \begin{subfigure}[b]{0.4\textwidth}
         \centering
         \includegraphics[width=\textwidth]{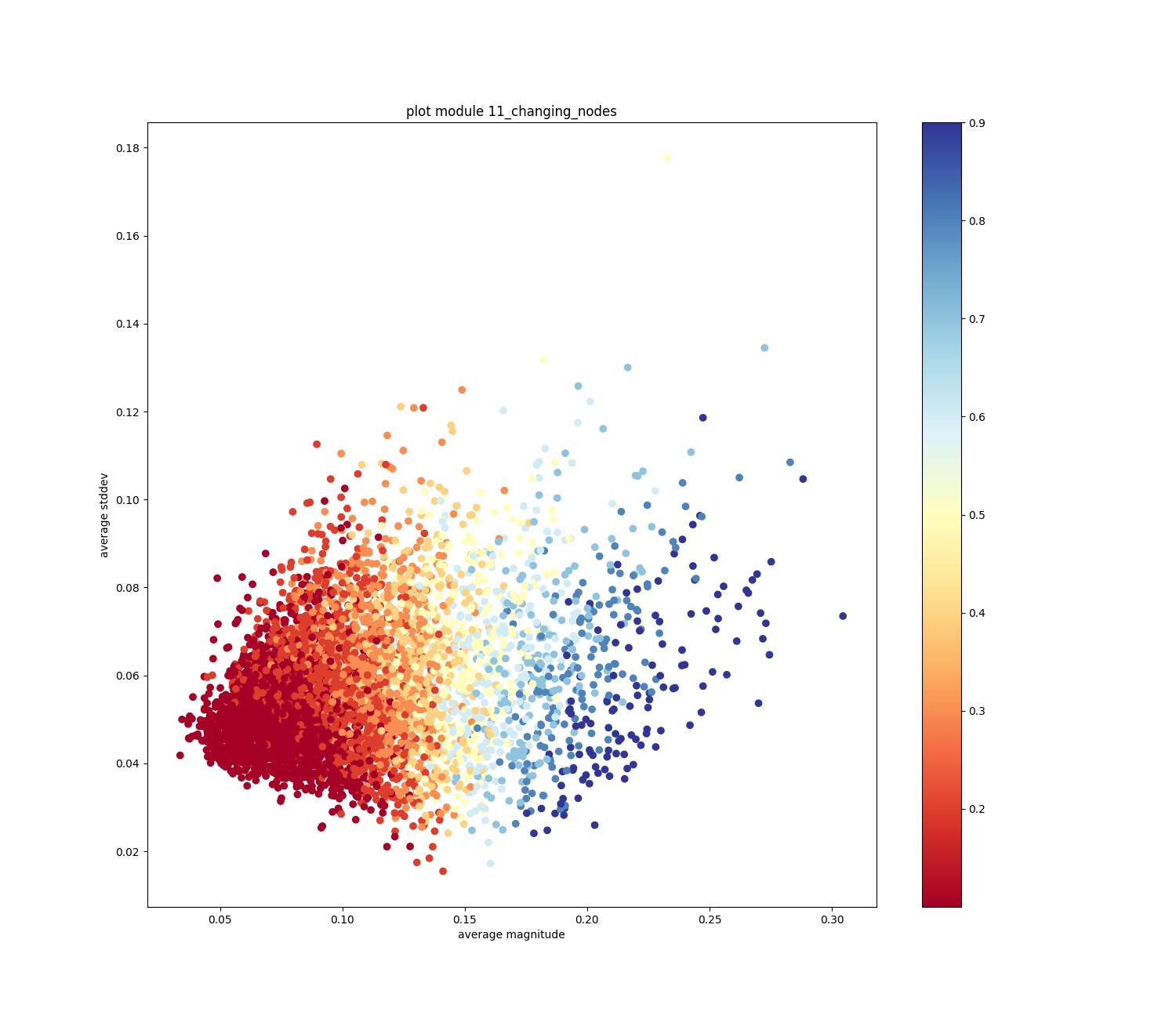}
     \end{subfigure}
     \begin{subfigure}[b]{0.4\textwidth}
         \centering
         \includegraphics[width=\textwidth]{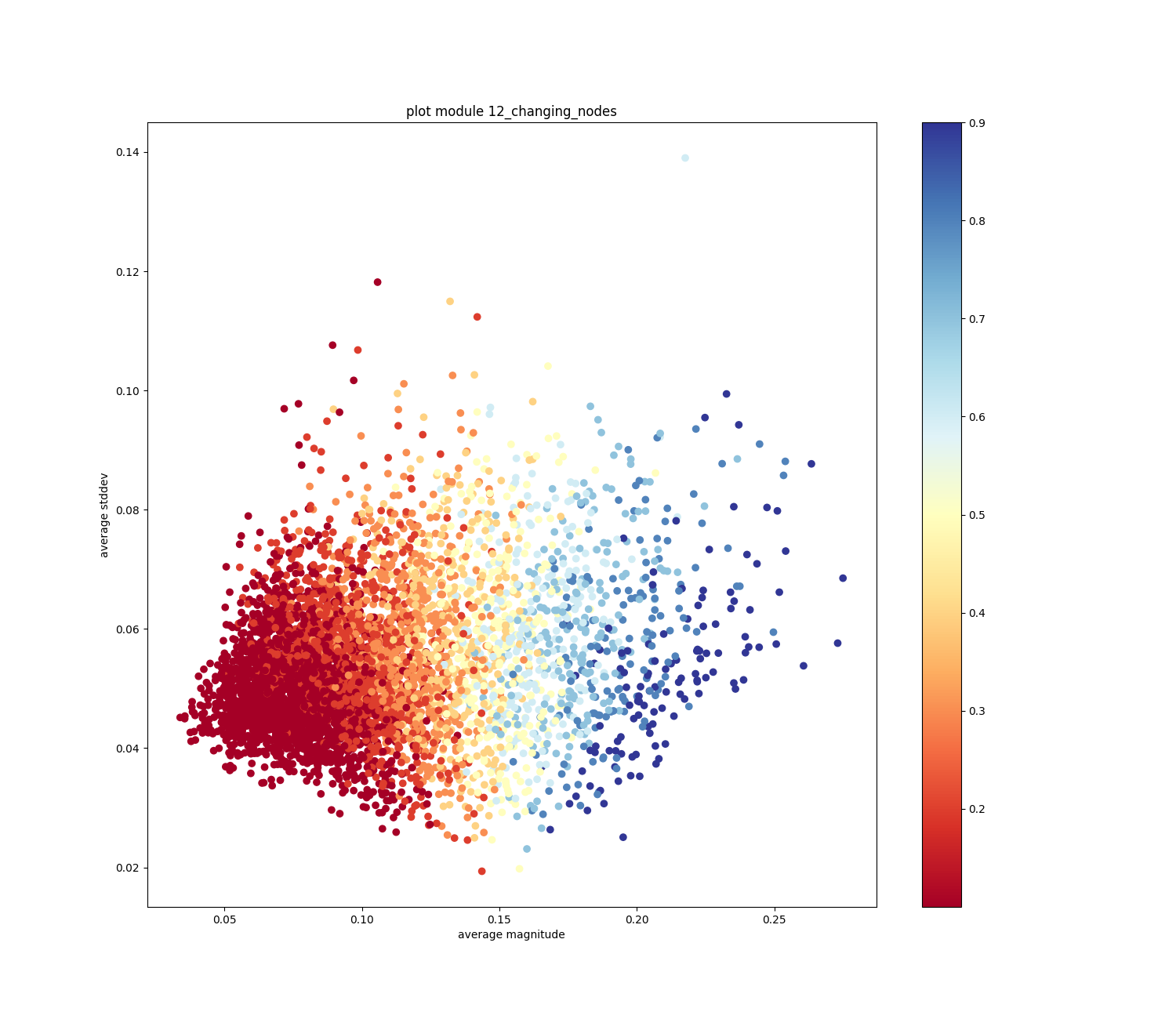}
     \end{subfigure}
     \begin{subfigure}[b]{0.4\textwidth}
         \centering
         \includegraphics[width=\textwidth]{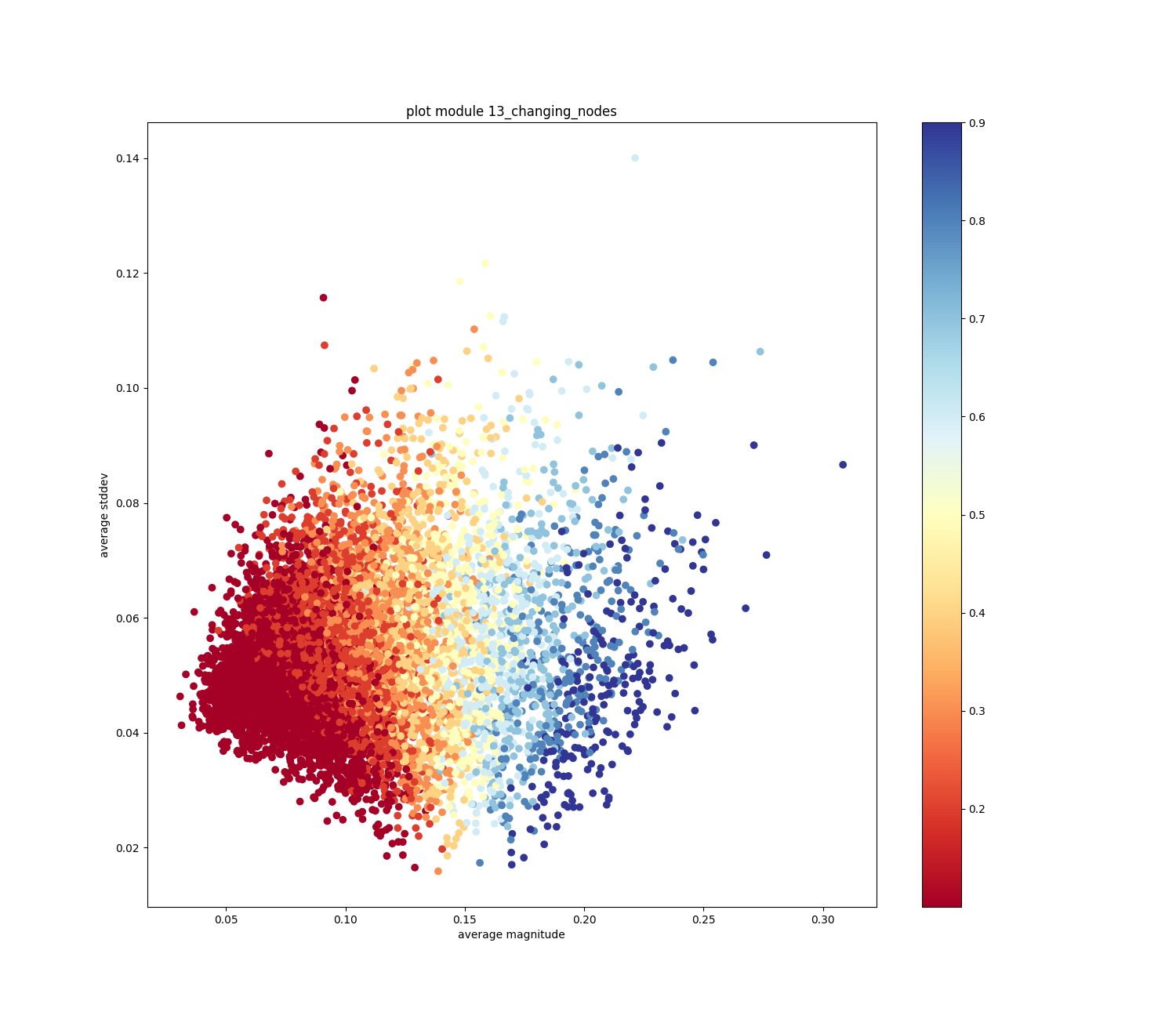}
     \end{subfigure}
    \caption{Plot showing the weights that change the mask (with Magnitude pruning) when data is reshuffled. X--axis is the average magnitude of the weight and Y--axis is its standard deviation. The architecture is ResNet20 trained on CIFAR-10. (layers 9-15).}
    \label{fig:changing_weights_CIFAR10_9_15}
\end{figure*}
\newpage

\end{document}